\documentclass[letterpaper]{article} % DO NOT CHANGE THIS
\usepackage{oplotsymbl}

\usepackage{caption}
\usepackage{subcaption}
\usepackage{adjustbox}
\usepackage{array}
\usepackage{algorithm}
\usepackage{algorithmicx}
\usepackage{algpseudocode}
\usepackage{multirow}
\newcolumntype{Z}[0]{>{\centering\arraybackslash}X}%
\newcolumntype{P}[0]{>{\hsize=.4\hsize}Z}%

\usepackage{amsmath}
\usepackage{amssymb}
\usepackage{mathtools}
\usepackage{amsthm}
\usepackage{xcolor} 

\usepackage[capitalize,noabbrev]{cleveref}

\theoremstyle{plain}

\theoremstyle{definition}

\theoremstyle{remark}

\usepackage{aaai23}  % DO NOT CHANGE THIS
\usepackage{times}  % DO NOT CHANGE THIS
\usepackage{helvet}  % DO NOT CHANGE THIS
\usepackage{courier}  % DO NOT CHANGE THIS
\usepackage[hyphens]{url}  % DO NOT CHANGE THIS
\usepackage{graphicx} % DO NOT CHANGE THIS
\usepackage{natbib}  % DO NOT CHANGE THIS AND DO NOT ADD ANY OPTIONS TO IT
\usepackage{caption} % DO NOT CHANGE THIS AND DO NOT ADD ANY OPTIONS TO IT
\usepackage{newfloat}
\usepackage{listings}
\DeclareCaptionStyle{ruled}{labelfont=normalfont,labelsep=colon,strut=off} % DO NOT CHANGE THIS
\floatstyle{ruled}
\newfloat{listing}{tb}{lst}{}
\floatname{listing}{Listing}
\title{A Tale of Two Latent Flows:
Learning Latent Space Normalizing Flow with Short-run Langevin Flow for Approximate Inference}
\author{
  Jianwen Xie,
  Yaxuan Zhu,
  Yifei Xu,
  Dingcheng Li,
  Ping Li\\
}
\affiliations{
    Cognitive Computing Lab\\
    Baidu Research\\
    10900 NE 8th St. Bellevue, WA 98004, USA\\
   \ \{jianwen.kenny,  yaxuanzhu.alvin, fei960922, dingchengl, pingli98\}@gmail.com
}

\usepackage{bibentry}
\begin{document}

\maketitle

\begin{abstract}
We study a normalizing flow in the latent space of a top-down generator model, in which the normalizing flow model plays the role of the informative prior model of the generator. We propose to jointly learn the latent space normalizing flow prior model and the top-down generator model by a Markov chain Monte Carlo (MCMC)-based maximum likelihood algorithm, where a short-run Langevin sampling from the intractable posterior distribution is performed to infer the latent variables for each observed example, so that the parameters of the normalizing flow prior and the generator can be updated with the inferred latent variables. We show that, under the scenario of non-convergent short-run MCMC, the finite step Langevin dynamics is a flow-like approximate inference model and the learning objective actually follows the perturbation of the maximum likelihood estimation (MLE). We further point out that the learning framework seeks to (i) match the latent space normalizing flow and the aggregated posterior produced by the short-run Langevin flow, and (ii) bias the model from MLE such that the short-run Langevin flow inference is close to the true posterior. Empirical results of extensive experiments validate the effectiveness of the proposed latent space normalizing flow model in the tasks of image generation, image reconstruction, anomaly detection, supervised image inpainting and unsupervised image recovery.
\end{abstract}

\section{Introduction}

A flow-based model~\cite{DinhKB14,dinh2016density,KingmaD18}, also known as normalizing flow, represents a data distribution by a non-linear invertible transformation of a simple distribution, e.g., Gaussian distribution. Thanks to the special design of the computation layers~of~the transformation, the flow-based model defines a
normalized probability distribution explicitly, making it convenient for learning, sampling and inference. Specifically, its learning can be handily accomplished by the maximum likelihood estimation with the tractable density function, its sampling can be efficiently achieved by a direct mapping of samples drawn from a Gaussian distribution, and its inference can be easily realized by the inverse transformation of data. Therefore, flow-based~models~have been widely used for direct modeling of various types of data, such as images~\cite{KingmaD18}, videos~\cite{KumarBEFLDK20}, point clouds~\cite{yang2019pointflow}, audio~\cite{PingPZ020} etc.  

In spite of the computational convenience, the expressivity of a normalizing flow is constrained by the unnatural design of the invertible transformation. Increasing the model expressivity typically requires significantly enlarging the network structure, thus leading to difficulty in constructing the network architecture and optimizing the network parameters. In order to foster the strengths and circumvent the shortcomings of the normalizing flow, we propose to learn a normalizing flow in the low-dimensional latent space of observed data. The resulting generative model becomes a normalizing flow standing on top of a top-down generator network. The latent space normalizing flow model can not only reserve the computational merit, but also focus on capturing regularities and rules of the data in the continuous latent space without having to care about the data type and data complexity. % in designing the architecture of the flow-based model.    
%is able to effectively acquire the regularities and rules in the data since the flow-based model is built on top of an expressive top-down network. 
Also, the learning and the design of a normalizing flow in the latent space is much easier than those in the data space due to the low dimensionality of the latent space. Typically, it can be parameterized by a small invertible multi-layer perceptron that consists of several layers.

Besides, the proposed latent space normalizing flow can be considered an informative prior distribution of the latent variables in the top-down generator model. Compared to the commonly used isotropic Gaussian (or uniform) non-informative prior in the conventional generator model, the informative prior represented by a trainable normalizing flow can significantly increase the expressive power of the entire generator. %More importantly, the ease of sampling and the tractability of probability density of the flow-based model make it a good fit for the role of  prior distribution in a deep latent variable framework in the sense that the flow-based prior can be jointly trained with the top-down network via maximum likelihood estimation and the data generation from the learned model can be explicit and fast.   
The normalizing flow model is a good fit for the role of prior distribution in the deep latent variable framework in the sense that (i) the normalizing-flow-based prior can be jointly trained with the top-down network due to the tractability of the prior probability density and (ii) the data generation from the learned model can be explicit and fast because of the ease of the prior sampling. We call our model the latent space flow-based prior model (LFBM).   

In this work, we propose a novel and principled way to jointly train the flow-based prior model and the generator via maximum likelihood estimation (MLE). The training process alternates the inference step and the learning step. Each inference step involves Markov chain Monte Carlo (MCMC) sampling of the posterior distribution, which is guided by both the flow-based prior distribution and the top-down generator, for the sake of inferring latent variables of each observed example, while, in each learning step, the parameters of the flow-based prior is updated by maximizing the likelihood of the inferred latent variables and the parameters of the generator are updated by minimizing the regression error between the inferred latent variables and the corresponding observed examples. %We call this framework the LFBM-MCMC. 
As to the MCMC inference, we adopt a short-run Langevin dynamics, i.e., a finite-step Langevin dynamics initialized with a fixed Gaussian distribution. Even though the short-run MCMC is an approximate inference due to the non-convergence, the normalizing flow prior can be adjusted based on the approximately inferred latent variables to mend the discrepancy between the aggregate posterior distribution and the prior distribution. We show that the learning algorithm based on short-run MCMC inference is a perturbation of MLE. In contrast to the latent space energy-based prior model (LEBM)~\cite{NijkampP0ZZW20}, in which short-run MCMCs are applied to not only the posterior sampling but also the EBM prior sampling, our model is able to perform an exact prior sampling via ancestral sampling and thus less biased from MLE than the LEBM.
%our LFBM-MCMC is able to perform an exact prior sampling via ancestral sampling and thus less biased from MLE and more computationally efficient than the LEBM. 
Compared to the baseline that trains the same LFBM along with an extra bottom-up inference model (with additional parameters) for variational auto-encoders (VAEs) ~\cite{KingmaW13}, our framework with a Langevin flow for approximate inference exhibits superiorities, e.g., better empirical performance, less model parameters, and free of inference network design. %Regardless of network architecture, deep latent variable models differ in their selections of the prior model and the inference process. 
Table \ref{tab:propertyanalysis} presents an analytical comparison of different latent variable models using various combinations of prior and posterior components. This comparison can clearly show the advantage of our proposed model and the motivation of this work. Specifically, this paper is motivated by proposing a practical, design-friendly and lightweighted latent variable framework with informative prior.

We summarize the main contributions of our paper as follows: (1) We propose a simple but novel deep generative model where a latent space normalizing flow, which serves as a prior, stands on a single top-down generator network. (2) We propose a natural and principled maximum likelihood learning algorithm to jointly train the latent space normalizing flow and the top-down network with MCMC-based inference over latent variables. (3) We also propose to use short-run Langevin flow as an approximate inference for efficient training. (4) We provide theoretical understanding of the proposed learning framework.  (5) We provide extensive and strong experimental results on different aspects, including image synthesis, inference, reconstruction, inpainting and recovery, to validate the effectiveness of the proposed models and learning algorithms.

\begin{table*}[ht]
\centering
\begin{tabular}{l|ccccccc}
\multirow{2}{*}{Methods (prior + posterior)}   & easy   & informative & fast      & fast      & less       & fast   & practical  \\
                          & design & prior       & sample & inference & parameters & training \\ \hline
Normalizing Flow + short-run MCMC       &    \circletfill    &      \circletfill       &     \circletfill      &     \circletfillhl      &     \circletfill       &   \circletfillhl     & \circletfill \\
Normalizing Flow + long-run MCMC        &   \circletfill    & \circletfill            &    \circletfill       &      \circlet     &     \circletfill        &  \circlet      &  \circlet  \\
Normalizing Flow + inference net        &   \circlet   &   \circletfill          &     \circletfill      &       \circletfill     &       \circlet      &    \circletfill   &  \circletfill \\
EBM + short-run MCMC      &   \circletfill     &  \circletfill           &   \circlet        &         \circletfillhl   &        \circletfill     &  \circletfillhl    &  \circletfill   \\
EBM + long-run MCMC       &     \circletfill    &  \circletfill           &    \circlet       &   \circlet        &      \circletfill       &    \circlet  &   \circlet  \\
EBM + inference net       &    \circlet     &    \circletfill         &    \circlet       &    \circletfill       &     \circlet        &  \circletfill   &  \circletfill   \\
Gaussian + short-run MCMC &    \circletfill     &  \circlet           &     \circletfill      &  \circletfillhl         &     \circletfill        &      \circletfillhl  &  \circletfill \\
Gaussian + long-run MCMC  &   \circletfill      &  \circlet           &    \circletfill       &    \circlet       &     \circletfill        &   \circlet   &    \circlet \\
Gaussian + inference net  &    \circlet     &   \circlet          &     \circletfill      &  \circletfill         &     \circlet        &   \circletfill   &   \circletfill  \\  
\end{tabular}
\caption{A comparison of latent variable models with different priors and inference process. ($\circletfill$ is good, $\circletfillhl$ is OK, and $\circlet$ is bad.) } 
\label{tab:propertyanalysis}
\end{table*}

\section{Related Work}

\paragraph{Data Space Flow-based Models and Latent Space Flow-based Models.} This work builds upon the techniques of normalizing flows, which have been widely applied to data space for representing high-dimensional data distributions directly~\cite{DinhKB14,dinh2016density,KingmaD18,yang2019pointflow,PingPZ020,KumarBEFLDK20,ShiXZZZT20}. This work moves the flow-based models from data space to latent space for representing low-dimensional latent manifolds from which the observed data are generated. Some prior works, such as~\citet{0022KSDDSSA17,huang2017learnable,MaZH19,xiao2019generative,DingG21}, study flow-based priors in VAE scenarios. Our paper mainly focus on maximum likelihood learning of flow-based priors with short-run Langevin MCMC~inference. 

\paragraph{Top-down Models with Implicit Density and Explicit Density.} A top-down model is a nonlinear generalization of the factor analysis model~\cite{rubin1982algorithms} and its density is generally implicit and intractable. Likelihood-based training of a top-down generator model typically relies on either the Markov chain Monte Carlo sampling that evaluates the posterior distribution, such as~\citet{HanLZW17,XieGZZW19} or the variational inference that approximates the posterior distribution, such as~\citet{KingmaW13}. A normalizing flow is a special top-down model that defines an explicit and normalized probability density and it can be learned by maximizing the exact likelihood of the observed data through a gradient ascent algorithm. Our model also belongs to a top-down model that stacks a top-down model with explicit density on top of the other one with implicit density. Similar to other generic top-down generator models, sampling from our model can also be achieved via direct ancestral sampling. However, the flow-based model at the top serves as a flexible and tractable prior distribution for the top-down model at the~bottom.

\paragraph{Latent Space Energy-based Prior.} Recently,~\citet{Pang0NZW20,PangZ0W21,PangW21,nipsZhang21} propose to use deep energy-based models (EBM) \cite{XieLZW16} to serve as flexible priors for top-down generation models. Because an EBM prior defines an unnormalized density, drawing samples from such a prior requires iterative MCMC sampling, thus bringing in inconveniences during the training and sampling of the whole model. Given the inferred latent variables, the update of the latent space EBM prior involves a non-convergent short-run MCMC that samples from the prior, which leads to a biased estimation of the EBM prior distribution \cite{NijkampHZW19,coopflow}. In contrast, our normalizing-flow-based prior model doesn't have such an issue due to its properties of tractable density and fast generation.      

\paragraph{MCMC-based Inference.}~Our~LFBM~is~related~to MLE of top-down generator model with MCMC-based inference, which is also called the alternating back-propagation (ABP) learning in~\citet{HanLZW17,XieGZZW19,XieGZZW20,ZhuXLE19,ZhangXB20,AnXL21,XingGHZW22}. The original ABP algorithm is designed to train a generator network with a known and simple Gaussian prior. The LEBM~\cite{Pang0NZW20} generalizes the ABP algorithm to a framework with a trainable and unnormalized energy-based prior model, while our paper generalizes the ABP algorithm to a framework with a trainable and normalized flow-based prior model.  

\paragraph{Variational Inference.} 
Most regular variational auto-encoder (VAE) frameworks employ a fixed Gaussian distribution as the prior and use reparameterization trick~\cite{KingmaW13} to design the amortized inference network that outputs a Gaussian posterior. Some prior works, such as~\citet{RezendeM15,kingma2016improved}, improve the amortized inference network by using a normalizing flow that outputs a flexible posterior. Our LFBM parameterizes the prior by a normalizing flow, treats the built-in short-run Langevin MCMC as the inference model, and learns them simultaneously from the observed data by following the empirical Bayes principle.

\section{Proposed Framework}
\subsection{Normalizing Flow in Latent Space}
Let $x\in \mathbb{R}^D$ be an observed signal such as an image, and let $z \in \mathbb{R}^d$ be the latent variables of $x$. The joint distribution of the signal and the latent variables $(x,z)$ is given by $p_{\theta}(x,z)=p_{\alpha}(z)p_{\beta}(x|z)$, where the distribution of $z$, i.e., $p_{\alpha}(z)$, is the prior model parameterized by $\alpha$, and the conditional distribution of $x$ given $z$, i.e., $p_{\beta}(x|z)$, is the top-down generation model parameterized by $\beta$. For notational convenience, let $\theta=(\alpha,\beta)$.   

The top-down generaton model is a non-linear transformation of the latent variables $z$ to generate the signal $x$, in which the transformation is parameterized by a neural network $g_{\beta}: \mathbb{R}^d \rightarrow \mathbb{R}^D$, which is similar to the decoder of a variational auto-encoder (VAE)~\cite{KingmaW13}. To be specific, $x=g_{\beta}(z)+\epsilon$, where $\epsilon \sim \mathcal{N}(0,\sigma^2I_D)$ is an observation residual. Thus, $p_{\beta}(x|z)=\mathcal{N}(g_{\beta}(z), \sigma^2I_D)$, where the standard deviation $\sigma$ is a hyper-parameter and assumed to be given. 
 
For most existing top-down generative models, the prior models are assumed to be non-informative isotropic Gaussian distributions. In this paper, we propose to formulate the prior $p_{\alpha}(z)$ as a flow-based model~\cite{dinh2016density, DinhKB14, KingmaD18}, which is of the form
\begin{eqnarray}
z_0 \sim q_0(z_0), \hspace{1mm} z =f_{\alpha}(z_0),  
\label{eq:prior_gen}
\end{eqnarray}
where $q_{0}(z_0)$ is a base distribution and has a simple and tractable density, such as a Gaussian white noise distribution: $q_0(z_0)=\mathcal{N}(0,I_d)$. $f_{\alpha}: \mathbb{R}^d \rightarrow \mathbb{R}^d  $ is an invertible or bijective function, which is a composition of a sequence of invertible transformations, i.e., $f_{\alpha}(z_0) = f_{\alpha_L} \circ  \cdots \circ f_{\alpha_2} \circ f_{\alpha_1}(z_0)$, whose inverse and logarithm of the determinants of the Jacobians can be explicitly obtained in closed form. Examples of such architectures include NICE~\cite{dinh2016density}, RealNVP~\cite{DinhKB14}, and Glow~\cite{KingmaD18}.

For $l=1,...,L$, let $z_{l}=f_{\alpha_l}(z_{l-1})$ be a sequence of random variables transformed from $z_{0}$, and define $z:=z_L$. According to the change-of-variable law of probabilities, $q_{0}(z_0)dz_0 = p_{\alpha}(z)dz$, the density of the flow-based prior model can be written as
\begin{eqnarray}
p_{\alpha}(z)&=q_{0}(z_0) \frac{dz_0}{dz}= q_{0}(f_{\alpha}^{-1}(z))\left|\det \left(\frac{\partial f_{\alpha}^{-1}(z)}{\partial z}\right)\right| \nonumber\\
&= q_{0}(f_{\alpha}^{-1}(z)) \prod_{l=1}^L\left|\det \left(\frac{\partial z_{l-1}}{\partial z_l}\right)\right|,
\label{eq:prior_den}
\end{eqnarray} 
where $f_{\alpha}^{-1}(z) =f_{\alpha_1}^{-1}  \circ \cdots \circ f_{\alpha_{L-1}}^{-1} \circ f_{\alpha_L}^{-1}(z) $, and the determinant of the Jacobian matrix $(\partial z_{l-1}/\partial z_l)$ can be easy to compute with well-designed transformation functions in the flow-based models. Typically, choosing transformations that can lead to a triangle Jacobian matrix will simplify the computation of the determinant: $\left|\det  ( \partial z_{l-1} /\partial z_l  )\right|=\prod |\text{diag}(\partial z_{l-1} /\partial z_l)|$, where $\text{diag}()$ takes the diagonal elements of the Jacobian matrix. Thus, the flow-based prior model has two nice properties: (i) analytically tractable normalized density (i.e., Eq. (\ref{eq:prior_den})),  and (ii) easy to draw samples from by using ancestral sampling (i.e., Eq. (\ref{eq:prior_gen})).    

The flow-based prior model in Eq.~(\ref{eq:prior_gen}) can be regarded as a flow-based refinement or calibration of the original Gaussian prior distribution $q_{0}$, which is a widely used prior distribution in most top-down generative frameworks. There are two advantages of using flow-based prior models for generative learning: (i) This enables us to learn a flexible prior from data to capture more meaningful latent components compared to those using a simple Gaussian prior model; (ii) The original generator learns $g_{\beta}$ to map from a fixed unimodal prior $q_0$ to the highly multi-modal data distribution, while in the proposed framework, the flow-based prior model corrects $q_0$ such that $g_{\beta}$ can be easier to transform the calibrated distribution $p_{\alpha}$ to the data distribution.

\subsection{Maximum Likelihood Learning with Langevin Inference} 
The top-down generative model with a flow-based prior can be trained via maximum likelihood estimation. For the training examples $\{x_i,i=1,...,N\}$, the observed-data log-likelihood function is given by
\begin{equation}
    \mathcal{L}(\theta)=\frac{1}{N}\sum_{i=1}^{N} \log p_{\theta}(x_i), \label{eq:mle}
\end{equation}
where the marginal distribution is obtained by integrating out the latent variables $z$: $p_{\theta}(x)=\int p_{\theta}(x,z)dz =\int p_{\alpha}(z)p_{\beta}(x|z)dz$. Maximizing the log-likelihood function $\mathcal{L}(\theta)$ is equivalent to minimizing the Kullback-Leibler (KL) divergence between the model $p_{\theta}(x)$ and the data distribution $p_{\text{data}}(x)$. The gradient of $\mathcal{L}(\theta)$ is computed according~to
\begin{eqnarray}
\begin{aligned}
    \nabla_{\theta}& \log p_{\theta}(x) = \mathbb{E}_{p_{\theta}(z|x)} \left[\nabla_{\theta} \log p_{\theta}(x,z)\right]  \\     &=\mathbb{E}_{p_{\theta}(z|x)}[\nabla_{\theta} (\log p_{\alpha}(z)+  \log p_{\beta}(x|z))], \label{eq:gradient}
    \end{aligned}
\end{eqnarray}
where the posterior distribution of $z$ is given by $p_{\theta}(z|x)=p_{\theta}(x,z)/p_{\theta}(x) \propto p_{\alpha}(z)p_{\beta}(x|z)$. The inference distribution $p_{\theta}(z|x)$ is dependent on both the prior model $\alpha$ and the generation model $\beta$. %See Supplementary Material for a detailed derivation of Eq.(\ref{eq:gradient}).

For the flow-based prior model, 
\begin{eqnarray}
\begin{aligned}
    \log p_{\alpha}(z) &= \log q_{0}(f_{\alpha}^{-1}(z)) + \sum_{l=1}^{L} \log |\text{det} (\frac{\partial z_{l-1}}{\partial z_l})|  \\
    &= \log q_{0}(z_0) + \sum_{l=1}^{L} \text{sum}(\log |\text{diag} (\frac{\partial z_{l-1}}{\partial z_l})|) 
    \end{aligned}
\end{eqnarray}
where $\log()$ takes element-wise logarithm, and $\text{sum}()$ takes the sum over all elements in a vector. Given a datapoint $z$, computing its log-likelihood only need one  pass of the inverse function $f_{\alpha}^{-1}$. We define $l_{\alpha}(z)=\log p_{\alpha}(z)$ to explicitly indicate that computing the log-likelihood of $z$ under the flow-based model is computationally tractable and can be regarded as a function of $z$. The learning gradient of $\alpha$ for a datapoint $x$ is 
\begin{eqnarray}
\begin{aligned}
    \nabla_{\alpha} \log p_{\theta}(x)&= \mathbb{E}_{p_{\theta}(z|x)}[\nabla_{\alpha} \log p_{\alpha}(z)] \\
    &=\mathbb{E}_{p_{\theta}(z|x)}[\nabla_{\alpha} l_{\alpha}(z)]. \label{eq:gradient_prior}
    \end{aligned}
\end{eqnarray}
The update of the latent space flow-based prior model depends on the observed example $x$, but different from the original data space flow-based model, it treats the latent variables inferred from $x$ as a training example and seeks to maximize the log-likelihood of the inferred latent variables. The updated $p_{\alpha}(z)$ will further influence the inference accuracy of $p_{\theta}(z|x)$.   

As to the generation model, the learning gradient of $\beta$ for a datapoint $x$ is
\begin{eqnarray}
    \nabla_{\beta} \log p_{\theta}(x)= \mathbb{E}_{p_{\theta}(z|x)}[\nabla_{\beta} \log p_{\beta}(x|z)]. \label{eq:gradient_generation}
\end{eqnarray}
Since $p_{\beta}(x|z)$ is in the form of a Gaussian distribution with a mean of $g_{\beta}(z)$ and a standard deviation of $\sigma$, $\nabla_{\beta} \log p_{\beta}(x|z)= \nabla_{\beta} (-\frac{1}{2\sigma^2}||x-g_{\beta}(z)||^2+\text{const})= \frac{1}{\sigma^2}(x-g_{\beta}(z))\nabla_{\beta}g_{\beta}(z)$. 

The learning gradient of $\theta$ in Eq.(\ref{eq:gradient}) is decomposed into the learning gradient of $\alpha$ in Eq.(\ref{eq:gradient_prior}) and the learning gradient of $\beta$ in Eq.(\ref{eq:gradient_generation}), both of which involve the expectation with respect to the intractable posterior $p_{\theta}(z|x)$, which can be approximated by drawing samples from $p_{\theta}(z|x)$ and then computing the Monte Carlo average. Sampling from $p_{\theta}(z|x)$ can be achieved by Langevin dynamics that iterates
\begin{eqnarray}
    z_{(k+1)} = z_{(k)} + \xi  \nabla_{z} \log p_{\theta}(z_{(k)}|x)+\sqrt{2 \xi} \epsilon_{(k)}; \nonumber\\ 
     \hspace{2mm} z_{(0)}\sim q_{0}(z), \epsilon_{(k)} \sim \mathcal{N}(0,I_d),
    \label{eq:MCMC_post}
\end{eqnarray}
where $k$ indexes the Langevin time step, $\nabla_{z} \log p_{\theta}(z|x)=\nabla_{z} \log p_{\alpha}(z)+ \nabla_{z} \log p_{\beta}(x|z) = \nabla_{z} l_{\alpha}(z)+ \frac{1}{\sigma^2}(x-g_{\beta}(z))\nabla_{z}g_{\beta}(z)$, and $\xi$ is the Langevin step size. Algorithm~\ref{alg1} presents the learning algorithm of~LFBM. We use the Adam optimizer \cite{KingmaB14} to update the parameters.

\subsection{Perturbation of Maximum Likelihood}

The Langevin inference in Eq.(\ref{eq:MCMC_post}) is practically  non-mixing and non-convergent because MCMC chains from different stating points can get trapped in local modes of the posterior distribution or MCMC chains are not sufficiently long to converge. Using a short-run non-convergent MCMC, e.g., a finite-step Langevin dynamics, as a flow-like approximate inference is more computationally efficient than using a long-run MCMC and more implementationally convenient than using an amortized inference network. We use $\hat{p}_{\theta}(z|x)$ to denote the short-run Langevin flow distribution, which is obtained by running a $K$-step Langevin dynamics starting from a fixed initial distribution $q_0(z)$ toward the true posterior ${p}_{\theta}(z|x)$. 
Strictly speaking, Algorithm~\ref{alg1} with a short-run MCMC inference is a perturbation of maximum likelihood. That is, given $\theta_t$ at iteration $t$, the learning gradient of $\theta$ is
\begin{eqnarray}
\mathcal{\hat{L}}(\theta) = \mathcal{L}(\theta)- \frac{1}{N}\sum_{i=1}^{N} \mathbb{D}_{\text{KL}}(\hat{p}_{\theta_t}(z_i|x_i)||p_{\theta}(z_i|x_i)),
\label{eq:short_run_loss0}
\end{eqnarray}
which is a lower bound of  $\mathcal{L}(\theta)$. Theoretically, if $K~\rightarrow~\infty$, $\xi \rightarrow 0$, then $\mathbb{D}_{\text{KL}}(\hat{p}_{\theta}(z|x)||p_{\theta}(z|x)) \rightarrow 0$. The estimate of $\theta$ is an MLE solution. However, $\mathbb{D}_{\text{KL}}(\hat{p}_{\theta}(z|x)||p_{\theta}(z|x))\neq0$ in practise, the estimate of $\theta$ is biased from~MLE. 

In Eq.(\ref{eq:short_run_loss0}), the first term $\mathcal{L}(\theta)$ corresponds to MLE, i.e., we want $p_{\theta}(x)$ to be close to $p_{\text{data}}$, while the second KL-divergence term means that we want to bias $\theta$ from MLE so that $p_{\theta}(z|x)$ is close to $\hat{p}_{\theta_t}(z|x)$. In fact, this is not a bad thing because it means we intend to bias the model so that short-run MCMC is close to the true posterior.

The algorithm converges when learning gradients equal to zeros. The resulting estimators of $\theta=(\alpha,\beta)$ solve the following estimating equations: %(See Supplementary Material for more explanation): 
\begin{align}
& \frac{1}{N}\sum_{i=1}^{N} \mathbb{E}_{\hat{p}_{\theta}(z_i|x_i)}[\nabla_{ \alpha} \log p_{\alpha}(z_i)]=0, \\ %\hspace{4mm}
& \frac{1}{N}\sum_{i=1}^{N} \mathbb{E}_{\hat{p}_{\theta}(z_i|x_i)}[\nabla_{ \beta} \log p_{\beta}(x_i|z_i)]=0.
\end{align}

%\subsection{Comparison with unnormalized energy-based prior}
In contrast to LFBM, the latent space energy-based prior model (LEBM)~\cite{Pang0NZW20} adopts an unnormalized energy-based density $p^{\text{ebm}}_{\alpha}(z)\propto \exp(f_{\alpha}(z))q_0(z)$ as the prior. The MLE of LEBM requires not only MCMC inference from $p_{\beta}(z|x)$ but also MCMC sampling from $p^{\text{ebm}}_{\alpha}(z)$. The short-run MCMC inference and sampling cause the learning algorithm of the LEBM to be a perturbation of MLE. Given $\theta_t$, the update of $\theta$ is based on the gradient of 
\begin{eqnarray}
\mathcal{\hat{L}}_{\text{LEBM}}(\theta) &= \mathcal{L}(\theta)- \frac{1}{N}\sum_{i=1}^{N} \mathbb{D}_{\text{KL}}(\hat{p}_{\theta_t}(z_i|x_i)||p_{\theta}(z_i|x_i)) \nonumber\\
&+ \mathbb{D}_{\text{KL}}(\hat{p}^{\text{ebm}}_{\alpha_t}(z)||p^{\text{ebm}}_{\alpha}(z)), \label{eq:short_run_ebm_loss0}
\end{eqnarray}
where $\hat{p}^{\text{ebm}}_{\alpha}(z)$ represents the short-run MCMC distribution. The second KL-divergence term in Eq.(\ref{eq:short_run_ebm_loss0}) is due to imperfect sampling from the EBM prior. Our flow-based prior $p_{\alpha}(z)$ is capable of exact sampling, therefore the second KL-divergence term disappears in Eq.(\ref{eq:short_run_loss0}). Thus, $\mathcal{\hat{L}}_{\text{LEBM}}(\theta)$ is more biased from MLE than $\mathcal{\hat{L}}(\theta)$. 

\begin{algorithm}
%\small 
\caption{Maximum likelihood learning of latent space normalizing flow model}\label{alg1}
\textbf{Input}: (1) Observed signals for training $\{x_i\}_{i}^{N}$; (2) Maximal number of learning iterations $T$; (3) Numbers of Langevin steps for posterior $K$; (4) Langevin step size $\xi$ for the posterior; (5) Learning rates for flow-based prior model and the generation model $\{\eta_\alpha,\eta_\beta\}$.\\
\textbf{Output}: 
Parameters $\beta$ for the generation model and $\alpha$ for the flow-based prior model
\begin{algorithmic}[1]
\State Randomly initialize $\alpha$ and $\beta$ 
\For{$t \leftarrow  1$ to $T$}
\State Sample a batch of observed examples $\{x_i\}_i^n$
\State For each $x_i$, sample the posterior $z_i \sim p_{\theta}(z|x_i)$ using $K$ Langevin steps in  Eq.(\ref{eq:MCMC_post}) with a step size $\xi$. 
\State Update flow-based prior by Adam optimizer  with the gradient $\nabla{\alpha}$ in Eq.(\ref{eq:gradient_prior}) and a learning rate $\gamma_\alpha$.
\State Update generation model by Adam optimizer with the gradient $\nabla{\beta}$ in Eq.(\ref{eq:gradient_generation}) and a learning rate $\gamma_\beta$.
\EndFor
\end{algorithmic} 
% \vspace{5mm}
\end{algorithm}

\subsection{Matching Normalizing Flow Prior and Aggregated Langevin Flow Posterior}

We further reveal the interaction between the normalizing flow prior and the Langevin flow
posterior in latent space during maximum likelihood estimation. The maximum likelihood learning minimizes the KL-divergence between the aggregated posterior and the prior. %$p_{\{\text{data},\theta\}}(z) = \int p_{\text{data}}(x) p_{\theta}(z|x) dx$ is the aggregated posterior.  $p_{\{\text{data},\theta\}}(x|z) = p_{\text{data}}(x)p_{\theta}(z|x)/p_{\{\text{data},\theta\}}(z) $ is the generation model based on data distribution and the posterior distribution. 
Let $\tilde{p}(x,z)=p_{\text{data}}(x)p_{\theta}(z|x)$, and then the aggregated posterior is $\tilde{p}(z)=\int \tilde{p}(x,z) dx =\mathbb{E}_{p_{\text{data}}(x)}[p_{\theta}(z|x)]$. Also, we have $\tilde{p}(x,z)=\tilde{p}(z)\tilde{p}(x|z)$. 
We show that 
\begin{eqnarray}
\begin{aligned}
&\mathbb{D}_{\text{KL}}(p_{\text{data}}(x)||p_{\theta}(x))\\
=&\mathbb{D}_{\text{KL}}(p_{\text{data}}(x)p_{\theta}(z|x)||p_{\theta}(x)p_{\theta}(z|x))\\
=&\mathbb{D}_{\text{KL}}(\tilde{p}(z)\tilde{p}(x|z)||p_{\alpha}(z)p_{\beta}(x|z))\\
=&\mathbb{D}_{\text{KL}}(\tilde{p}(z)||p_{\alpha}(z)) + \mathbb{D}_{\text{KL}}(\tilde{p}(x|z)||p_{\beta}(x|z)),
\end{aligned}
\end{eqnarray}
which means that the maximum likelihood learning of the whole latent variable models involves a behavior that minimizes the KL divergence between the aggregated posterior and the normalizing flow in the latent space. 

The above analysis is based on the ideal scenario with a long-run convergent Langevin inference. In our framework, we assume and allow short-run Langevin sampling, and then our objective turns into maximizing the perturbation of maximum likelihood, i.e., $\log p_{\theta}(x) - \mathbb{D}_{\text{KL}}(\hat{p}_{\theta_t} (z|x)  ||  p_{\theta}(z|x) )$, or equivalently minimizing $\mathbb{D}_{\text{KL}}(p_{\text{data}}(x)||p_{\theta}(x)) + \mathbb{D}_{\text{KL}}(\hat{p}_{\theta_t} (z|x)  ||  p_{\theta}(z|x) )$.
We now generalize the above analysis to the short-run non-convergent Langevin flow scenario. We show that
\begin{eqnarray}
\begin{aligned}
&\mathbb{D}_{\text{KL}}(p_{\text{data}}(x) || p_\theta(x)) +\mathbb{D}_{\text{KL}}(\hat{p}_{\theta_t}(z|x) || p_\theta(z|x))\\
= &\mathbb{D}_{\text{KL}}(p_{\text{data}}(x) \hat{p}_{\theta_t}(z|x) || p_\theta(x) p_\theta(z|x))\\
=&\mathbb{D}_{\text{KL}}(p_{\text{data}}(x) \hat{p}_{\theta_t}(z|x) || p_\alpha(z) p_\beta(x|z)) \\
= &\mathbb{D}_{\text{KL}}(p_{\text{data}}(z) \hat{p}_{\theta_t}(x|z) || p_\alpha(z)p_\beta(x|z))\\
= &\mathbb{D}_{\text{KL}}(p_{\text{data}}(z) || p_\alpha(z)) + \mathbb{D}_{\text{KL}}(\hat{p}_{\theta_t}(x|z) || p_\beta(x|z)),
\end{aligned}
\end{eqnarray}
where $p_{\text{data}}(z) = \int p_{\text{data}}(x) \hat{p}_{\theta_t}(z|x) dx$ is the aggregated posterior.  $\hat{p}_{\theta_t}(x|z) = p_{\text{data}}(x)\hat{p}_{\theta_t}(z|x)/p_{\text{data}}(z) $ is the generation model based on data distribution and the inference model (i.e., short-run Langevin flow). KL-divergence between conditional distributions is understood to be averaged over the variable $z$ being conditioned upon. In the above, we update $p_\alpha(z)$ to be close to $p_{\text{data}}(z)$. We update $p_{\beta}(x|z)$ to be close to $\hat{p}_{\theta_t}(x|z)$. Both $p_{\text{data}}(z)$ and $\hat{p}_{\theta_t}(x|z)$ depends on the short-run Langevin flow $\hat{p}_{\theta_t}(z|x)$. 

Therefore, the learning of our model accomplishes two things: (1) Let the flow-based prior match the aggregated posterior produced by the short-run Langevin flow inference model. (2) Bias the model from MLE so that the short-run Langevin flow inference is close to the true posterior.

\begin{figure*}[ht]
\centering
\begin{subfigure}{.3\linewidth}
    \centering
    \includegraphics[width=0.95\textwidth]{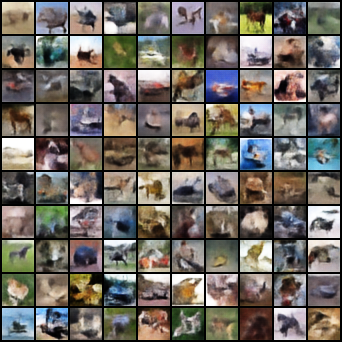}
    \caption{CIFAR10}
\end{subfigure}
    \hspace{-0.5em}
\begin{subfigure}{.3\linewidth}
    \centering
    \includegraphics[width=0.95\textwidth]{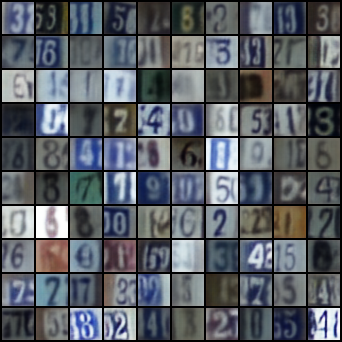}
    \caption{SVHN}
\end{subfigure}
  \hspace{-0.5em}
\begin{subfigure}{.3\linewidth}
    \centering
    \includegraphics[width=0.95\textwidth]{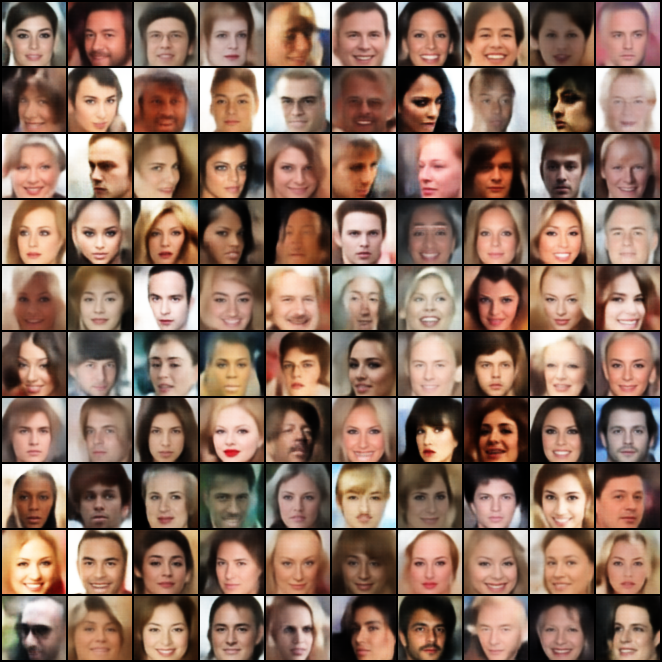}
    \caption{CelebA}
\end{subfigure}
\caption{Generated examples from the LFBM-MCMC models traind on the CIFAR10 (32$\times$ 32), SVHN, (32$\times$ 32) and CelebA (64$\times$ 64) datasets. The LFBM-MCMC model is trained with short-run Langevin flow as approximate inference.} 
\label{fig:sample_mcmc}
\end{figure*}

\section{Experiments}
\subsection{Experiment Settings} We conduct a set of experiments to examine the effectiveness of the proposed latent space flow-based prior model in image modeling. In particular, we evaluate the model from the perspectives of synthesis, reconstruction and inference. %Besides, we also analyze and compare the behaviors of the learning strategies that use MCMC-based inference (i.e., LFBM-MCMC) and amortized inference  (i.e., LFBM-VAE). 
We use SVHN~\cite{netzer2011reading}, CelebA~\cite{LiuLWT15}, and CIFAR-10~\cite{krizhevsky2009learning} image datasets for generation and reconstruction tasks, and use MNIST~\cite{mnist2010} dataset for anomaly detection. All the inpainting and recovery experiments are conducted on CelebA face dataset. The network architectures and the hyperparameters %of both LFBM-MCMC and LFBM-VAE 
are presented in Supplementary Material. The experiments are mainly for the sake of proof of concepts rather than state-of-the-art performance. We will compare our framework with two important baseline methods, which are (1) LFBM-VAE baseline, which trains LFBM using variational inference instead of MCMC inference, (2) LEBM baseline, which uses an energy-based model in the latent space. For fair comparison, the architecture designs of the top-down generators $g_{\beta}$ in LEBM, LFBM-VAE and ours are the same, and the  flow-based prior models $f_{\alpha}$ in LFBM-VAE and ours are also identical. The design of the inference network of LFBM-VAE follows \citet{kingma2016improved}, who parameterize the inference network by an encoder followed by an inverse autoregressive flow.  % as those in LFBM-VAE. %The architectures of the inference networks are exclusive for LFBM-VAE. 
The designed network architectures are expected to vary in different datasets due to various image sizes and pattern complexities. We use LFBM-MCMC to denote our LFBM using Langevin inference in order to distinguish from the baseline LFBM-VAE.

\subsection{Image Synthesis and Reconstruction} 
A well-trained latent variable generative model can be useful for generation and reconstruction. We train models on training images of SVHN, CelebA and CIFAR-10 respectively, and generate synthesized examples by first sampling latent vectors from the learned latent space normalizing flow and then transforming the vectors to image space. We calculate the Fréchet inception distance (FID)~\cite{HeuselRUNH17} to measure the quality of the synthesized images in Table~\ref{tab:compare}. 
We show generated samples of the learned LFBM-MCMC models in Figure~\ref{fig:sample_mcmc}.
The models can reconstruct images by first inferring the latent vectors from the images, and then mapping the inferred latent vectors back to data space. The inference of latent variables can be achieved by the MCMC in our LFBM-MCMC. The quality of reconstruction images are measured by mean squared error (MSE) in Table~\ref{tab:compare}. We mainly compare our methods with likelihood-based top-down models rather than adversarial frameworks. Specifically, we compare with traditional variational inference baselines, such as VAE~\cite{KingmaW13} and SRI~\cite{NijkampP0ZZW20} which use Gaussian priors and two VAE variants, 2sVAE and RAE, which learn their priors from posterior samples in the second stage. We also compare with MCMC inference frameworks, such as ABP~\cite{HanLZW17} and LEBM~\cite{Pang0NZW20} whose prior distributions are fixed Gaussian distribution and energy-based model, respectively. We also compare our LFBM-MCMC with the VAE variant, i.e., the LFBM-VAE. Our models outperform other baselines in terms of MSE and FID.

\begin{table*}[]
\begin{center}
\resizebox{.9\textwidth}{!}{
\begin{small}
\begin{tabular}{cc|ccccccc|cc}
\hline
\multicolumn{2}{c|}{\multirow{2}{*}{Models}}   & \multirow{2}{*}{VAE}    & \multirow{2}{*}{2sVAE}  & \multirow{2}{*}{RAE}  &  \multirow{2}{*}{SRI}   & \multirow{2}{*}{SRI (L=5)} & \multirow{2}{*}{ABP} & \multirow{2}{*}{LEBM}   & \multicolumn{2}{c}{LFBM}   \\ 
  & &     &   &   &    &  &  &   & VAE  & MCMC \\ \hline
%\multicolumn{2}{c|}{} & &~\cite{}&~\cite{}&~\cite{} & & &&(ours)&(ours)\\ \hline
%\multirow{2}{*}{MNIST}    & MSE & ???   & \\   & FID & \textbf{15.4494}     & \\ \hline 
\multirow{2}{*}{SVHN}    & MSE & 0.019  & 0.019  & 0.014 &  0.018 & 0.011  &  - & 0.008 &     0.005&\textbf{0.005}  \\ 
                         & FID & 46.78  & 42.81  & 40.02 &  44.86 & 35.23   & 49.71 & 29.44 & 24.96     &\textbf{23.64} \\ \hline 
 \multirow{2}{*}{Cifar10} & MSE & 0.057  & 0.056  & 0.027  &-     & -      & 0.018  & 0.020 & 0.020 & \textbf{0.016}
                          \\  
                          & FID & 106.37 & 72.90 & 74.16 &  -     & -       & 90.30 & 70.15 & 69.70  & \textbf{66.41}  \\ \hline
\multirow{2}{*}{CelebA}  & MSE & 0.021  & 0.021  & 0.018  & 0.020 & 0.015  &  - & 0.013 &   0.014  & \textbf{0.011} \\ 
                         & FID & 65.75  & 44.40  & 40.95  & 61.03 & 47.95   & 51.50 & 37.87 & 33.64     & \textbf{33.64} \\ \hline
\end{tabular}
\end{small}}
\end{center}
\caption{Quantitative results of image reconstruction and generation on different datasets. %The MSE and FID are used to evaluate test the reconstructed and generated images, respectively. 
}
\label{tab:compare}
\vspace{-0.1cm}
\end{table*}

\subsection{Supervised Image Inpainting} 
\label{sec:image_inpainting}
We can train an LFBM-MCMC from fully-observed training images, and then use the learned model to complete the missing pixels of testing images. Let $m$ be a matrix, which has the same number of dimension as that of an image $x$, with values ones indicating the visible pixels and zeros indicating the invisible ones (corrupted or occluded) of the image $x$, respectively. Suppose there is an incomplete image $x_{m}$ with missing pixels indicated by a mask $m$. With the learned flow-based prior model $p_{\alpha}(z)$ and the generation model $p_{\beta}(x|z)$, we can restore the missing pixels by first inferring the latent vectors $z$ from $x_{m}$, and then using $g_{\beta}(z_m)$ to generate an image $x$, which is a complete version of $x_m$. The inference can be easily performed by Langevin dynamics that follows Eq.(\ref{eq:MCMC_post}) with the gradient of a modified log-posterior $\nabla_{z} \log p_{\theta}(z|x_m,m)=\nabla_{z} \log p_{\alpha}(z)+ \nabla_{z} \log p_{\beta}(x_m|z,m) = \nabla_{z} l_{\alpha}(z)+ \frac{1}{\sigma^2}(m \odot (x_m-g_{\beta}(z)))\nabla_{z}g_{\beta}(z)$, where $\odot$ is the element-wise multiplication operator. That is, we evaluate the posterior over only the visible pixels of the images. Each testing image is occluded by a $40 \times 40$ region mask at the center. The resolution of the images are $64 \times 64$ pixels. For each testing image, our LFBM-MCMC model can generate diverse and meaningful inpainting results by using different $z_0$ sampled from $q_0(z)$ to initialize the Langevin chains for inference.

\begin{figure*}[t!]
\centering
\includegraphics[width=0.955\textwidth]{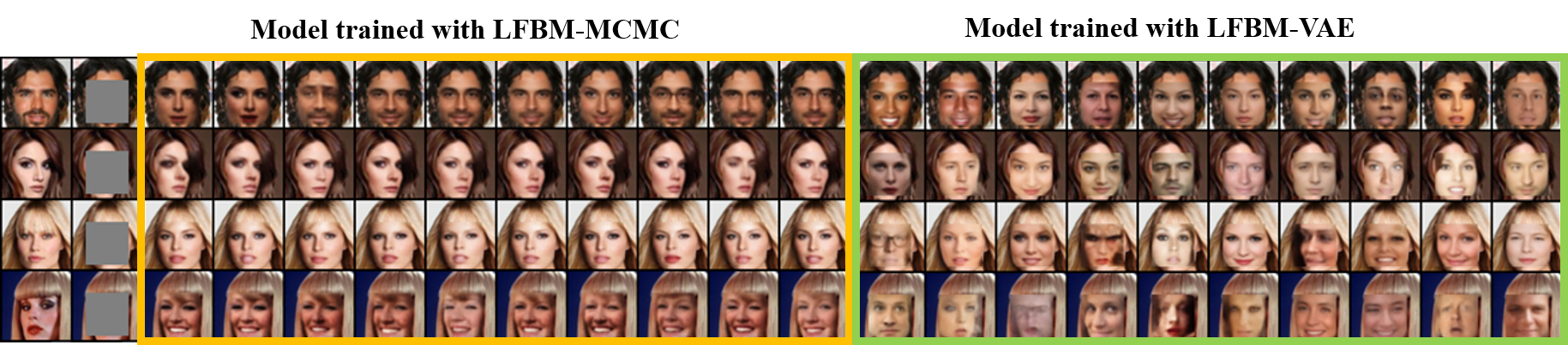}
\caption{Supervised image inpainting results on the CelebA dataset. Images in the first column are the original images. Images in the second column are the masked images to be inpainted. Images  in column 3 to column 12 (yellow panel) are inpainting results using the learned LFBM-MCMC model. Images in column 13 to column 22 (green panel) are inpainting results using the trained LFBM-VAE. For each panel, different columns correspond to different initializations of the inference process.}
\label{fig:pretrained_inpaint}
\end{figure*}

Figure \ref{fig:pretrained_inpaint} displays some qualitative results, where we compare our LFBM-MCMC model with the baseline LFBM-VAE. Both LFBM-MCMC and LFBM-VAE are trained on fully observed training images of CelebA dataset. For task of inpainting, given an incomplete image with a mask occluding pixels at the center, our LFBM-MCMC first infers the latent variables of the incomplete image by the Langevin flow, and then generate the occluded region of the image by the top-down generator taking the inferred variables as input. As to the LFBM-VAE, the learned inference model in the LFBM-VAE is hard to employ for inferring latent variables from an incomplete data, which means that even though the variational inference is computationally efficient, it is not suitable to infer latent variables from a partially observed image because the encoder can not map a portion of the image into the latent space. In order to use the LFBM-VAE for image inpainting, we abandon the learned inference network, and derive the Langevin dynamics from the learned prior and generator models in LFBM-VAE. We adopt the derived Langevin flow as in the LFBM-MCMC for inference and inpainting. As shown in Figure \ref{fig:pretrained_inpaint}, each row illustrates one inpainting task. The first column displays the original images, and the second column shows the testing images that need to be inpainted. The yellow panel shows inpainted images by the LFBM-MCMC, while the green panel shows inpainted images by the LFBM-VAE. Different columns in each panel show different inpainting results due to the different initialization (i.e., randomness) of the inference process. Although the LFBM-VAE adopts the MCMC inference for inpainting, its results are still not as good as those by our LFBM-MCMC. Some obvious artifacts are observed in the inpainting results of the LFBM-VAE. Further, we quantitatively evaluate the image inpainting performance of both LFBM-MCMC and LFBM-VAE in Table~\ref{tab:image_inpainting}. We use two metrics, i.e., FID and MSE, to measure the quality of the inpainting performance. We test the models in 10,000 images. For the MSE, we compute the per pixel difference between the inpianting result and the ground truth image within the masked region. Our model outperforms the baseline.

\begin{table}[h!]
\centering

\begin{tabular}{ccc}
\hline
Method & FID & MSE \\ \hline
LFBM-MCMC (ours) & \textbf{0.0557} & \textbf{0.1937}\\
LFBM-VAE  & 0.0562 & 0.2074 \\
%Gaussian  & ??? \\ 
\hline
\end{tabular}
\caption{Quantitative results of image inpainting.}
\label{tab:image_inpainting}
\end{table}

\subsection{Anomaly Detection} We evaluate our generative model on anomaly detection of MNIST~\cite{mnist2010} data. Given a latent variable model that is well-trained on normal examples, we can perform anomaly detection on a testing image $x$ by firstly inferring its latent variables $z$ and then computing the logarithm of the joint probability $\log p_{\theta}(x, z)=\log p_{\alpha}(z) + \log p_{\beta}(x|z)=l_{\alpha}(z)-\frac{1}{2\sigma^2}||x-g_{\beta}(z)||^2- \log \sigma \sqrt{2 \pi}$ as a decision score. The score should be high for a normal example and low for an anomalous one. We can see that a normal example can be well-reconstructed by the learned generator (i.e., small reconstruction error $||x-g_{\beta}(z)||^2$) with a correctly inferred latent variables (i.e., high log likelihood in the learned prior distribution $l_{\alpha}(z)$).

Following the same experiment setup as~\citet{zenati2018efficient,kumar2019maximum, Pang0NZW20}, we treat each class of digit images in MNIST dataset as anomaly examples and the remaining 9 classes of digit images as normal examples. We train models only on the normal examples, and then test the models on both the normal and anomalous examples. To quantify the model performance, we compute the area under the precision-recall curve (AUPRC) based on the decision function $\log p_{\theta}(x, z)$. We report the mean and variance of AUPRC scores over 10 runs of each experiment.We show the results of our LFBM in Table~\ref{tab:anormaly}, and compare with the related models, including the VAE~\cite{KingmaW13}, MEG~\cite{kumar2019maximum}, BiGAN-$\sigma$~\cite{zenati2018anomaldetection}, EBM-VAE~\cite{Han2020VAE-EBM}, LEBM~\cite{Pang0NZW20} and ABP model~\cite{HanLZW17}. From Table~\ref{tab:anormaly}, we can find that the proposed LFBM can obtain much better results than those of other~methods.

\begin{table*}[]
\begin{center}
\resizebox{1.02\textwidth}{!}{
\begin{small}
\begin{tabular}{c|ccccc}
\hline
Heldout Digit   & 1 & 4 & 5 & 7 & 9 \\  \hline 
VAE~\cite{KingmaW13}  & 0.063 & 0.337 & 0.325 & 0.148 & 0.104 \\
MEG ~\cite{kumar2019maximum} & 0.281 $\pm$ 0.035 & 0.401 $\pm$ 0.061 & 0.402 $\pm$ 0.062 & 0.290 $\pm$ 0.040 & 0.342 $\pm$ 0.034 \\ 
BiGAN-$\sigma$~\cite{zenati2018anomaldetection}   & 0.287 $\pm$ 0.023 & 0.443 $\pm$ 0.029 & 0.514 $\pm$ 0.029 & 0.347 $\pm$ 0.017 & 0.307 $\pm$ 0.028 \\ 

  EBM-VAE~\cite{Han2020VAE-EBM} & 0.297 $\pm$ 0.033 & 0.723 $\pm$ 0.042 & 0.676 $\pm$ 0.041 & 0.490 $\pm$ 0.041 & 0.383 $\pm$ 0.025 \\

LEBM ~\cite{Pang0NZW20} & 0.336 $\pm$ 0.008 & 0.630 $\pm$ 0.017 & 0.619 $\pm$ 0.013 & 0.463 $\pm$ 0.009 & 0.413 $\pm$ 0.010 \\ 
ABP~\cite{HanLZW17}  & 0.095 $\pm$ 0.028 & 0.138 $\pm$ 0.037 & 0.147 $\pm$ 0.026 & 0.138 $\pm$ 0.021 & 0.102 $\pm$ 0.033 \\ \hline
LFBM (ours)  & \bf{0.349 $\pm$ 0.002} & \bf{0.812 $\pm$ 0.007} & \bf{0.823 $\pm$ 0.009} & \bf{0.682 $\pm$ 0.004} & \bf{0.514 $\pm$ 0.008} \\  
%LFBM-VAE (ours) & 0.069$\pm$0.001 & 0.366$\pm$0.009 & 0.338 $\pm$ 0.007 & 0.139 $\pm$ 0.004 & 0.127 $\pm$ 0.001 \\ 
\hline
\end{tabular}
\end{small}}
\end{center}
\caption{AUPRC scores (larger is better) for unsupervised anomaly detection. 
%on the MNIST dataset. Results for our model are averaged over 10 experiments for variance
Results are averaged over 10 experiments.}
\label{tab:anormaly}
%\vspace{-0.5cm}
\end{table*}

\subsection{Analysis of Hyperparameters}
\begin{figure*}[h]
	\centering 
	\begin{subfigure}{.28\linewidth}
	\includegraphics[width=1\linewidth]{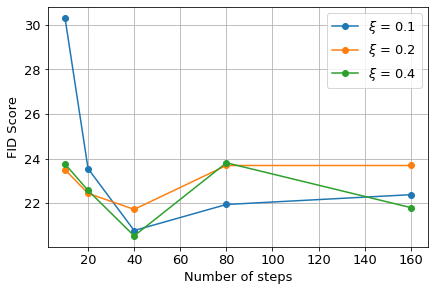}
	\caption{number of steps and step size} 
	\end{subfigure}
	\begin{subfigure}{.28\linewidth}
	\includegraphics[width=1\linewidth]{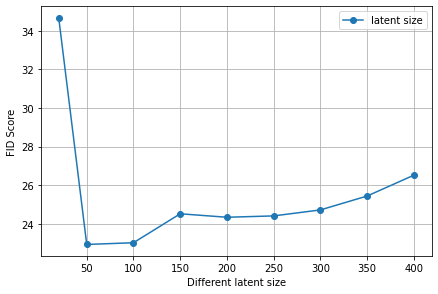}
	\caption{number of latent dimension} 
	\end{subfigure}
	\begin{subfigure}{.31\linewidth}
	\includegraphics[width=1\linewidth]{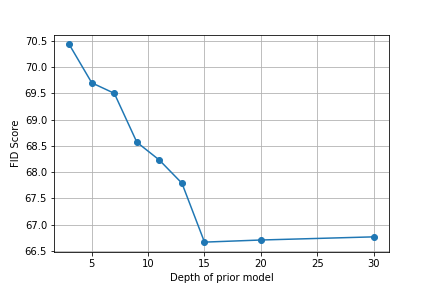}
	\caption{depth of normalizing flow} 
	\end{subfigure}
\caption{FIDs of LFBMs using different (a)  numbers of Langevin steps and step sizes, (b)  latent sizes, and (c) depths of the normalizing flow prior model. %Results are obtained on the SVHN dataset.
}
\label{ablation_study}
\end{figure*}

In Figure~\ref{ablation_study}(a), we show the effects of using different Langevin step sizes and numbers of steps for LFBM. We carry out experiments on SVHN dataset. We can see that the best FID score is achieved when the number of Langevin steps is around 40 and the model with 20 steps can already give us good results. On the other hand, the optimal step sizes for Langevin flows using different numbers of steps are different. Figure~\ref{ablation_study}(b) displays the effect of tuning latent size. We can see the best result can be achieved in the range from 50 to 100. Figure~\ref{ablation_study}(c) shows FID scores using different depths of the flow-based prior on Cifar10. We find that increasing the depth of the prior can improve the performance.

\subsection{Unsupervised Image Recovery}

\begin{figure*}[th]
    \centering
     \includegraphics[width=0.935\textwidth]{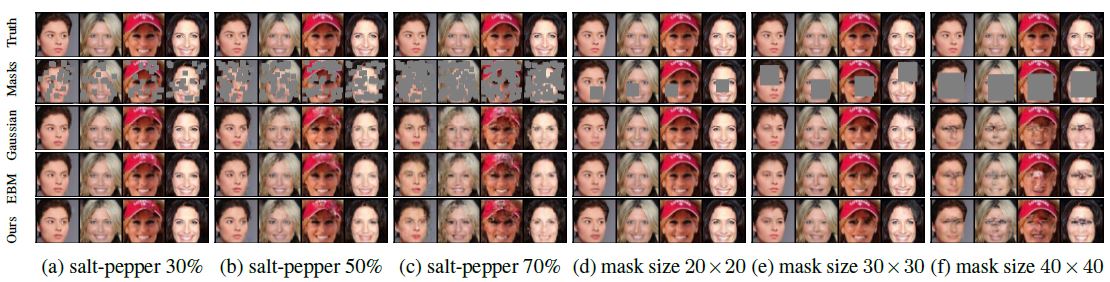}
     \caption{A comparison of unsupervised image recovery results by different methods on training images with different levels of occlusions. In each panel, the first row shows some original images that are used in the training process, the second row shows the corresponding occluded images with a certain occlusion level, and the third, fourth and fifth rows show the recovered images by models using Gaussian prior, EBM prior and normalizing flow prior, respectively.}
    \label{fig:rec_recovery}
\end{figure*}

\begin{figure*}[h!]
    \centering
     \includegraphics[width=0.93\textwidth]{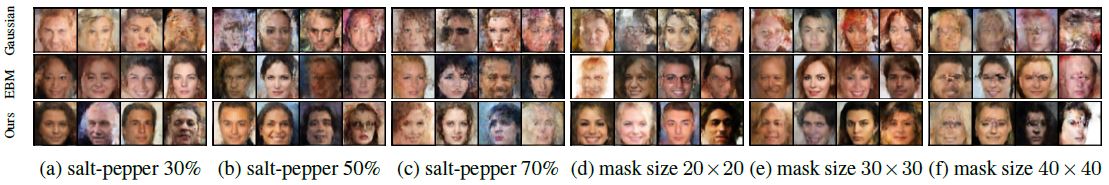}
    \caption{Image synthesis by models learned from incomplete images. Each panel represents a different level of occlusion.}
    \label{fig:recovery_gen}
\end{figure*}

The LFBM can be learned from incomplete training data, e.g., images with occluded pixels. The learning algorithm updates the model parameters by maximizing the likelihood of the visible pixels in training images. This also belongs to unsupervised image inpainting, which is essentially different from the task of supervised image inpainting shown in Section~\ref{sec:image_inpainting}. Traditional VAE-based frameworks, such as LFBM-VAE, are incapable of learning from incomplete data. To demonstrate this ability of the LFBM, we experiment on 10,000 occluded images that we create. They are selected from the CelebA dataset
 and occluded with different types of masks. A mask is randomly placed in each image. Two types of masks are designed, including single region mask and salt-and-pepper mask. We specify three different sizes of single regions, e.g., $20 \times 20$, $30 \times 30$, and $40 \times 40$, and three different occlusion percentages of salt-and-pepper
masks, e.g., $30\%$, $50\%$, and $70\%$. We compare our LFBM with a fixed Gaussian prior and an energy-based prior learned by MCMC inference. We can evaluate the performance from two aspects: (i) Recovery quality: The incomplete training images are gradually recovered as the learning algorithm proceeds. MSE between recovered images and the corresponding ground truth images at masked regions are calculated; (ii) Generation quality: We use FID to measure the visual quality of the synthesized images, which are generated by models learned from incomplete data. Table~\ref{tab:MSE_mask_recovery} and Table~\ref{tab:FID_mask_recovery} show comparisons of our method with the baselines using EBM prior and Gaussian prior in terms of MSE and FID, respectively. %The recovery images and generated images are shown in Supplementary Material.% The LFBM is more powerful than others。 % to learn from incomplete images.% in the sense that the synthesized images are~still~meaningful.
%\vspace{-2mm}
 Figure~\ref{fig:rec_recovery} shows the recovery results for learning from incomplete training images of the CelebA dataset. We further evaluate the generation capacities of the models trained in this scenario. We present qualitative results in Figure~\ref{fig:recovery_gen}, which shows the randomly generated images by the models learned from incomplete images with different occlusion levels. We can see that frameworks using Gaussian priors have the worst generated results. Our LFBM model using Langevin flow as an inference process can generate more realistic images than other baselines in the task of unsupervised learning.

\begin{table}[h!]
\centering
\begin{tabular}{cccc}
\hline
\multicolumn{4}{c}{Salt and pepper mask}                     \\ \hline
Occ \%    & 30\%           & 50\%           & 70\%           \\ \hline
flow (ours)      & \textbf{0.0244}  & \textbf{0.0317} & \textbf{0.0464} \\
EBM       & 0.0256          & 0.0319          & 0.0465          \\
Gaussian  & 0.0259         & 0.0326         & 0.0472         \\ \hline
\multicolumn{4}{c}{Single region mask}                       \\ \hline
mask size & $20 \times20$          & $30\times 30$          & $40 \times40$          \\ \hline
flow (ours)      & 0.0420 & 0.0587 & \textbf{0.0864} \\
EBM       & 0.0429         & 0.0684          & 0.0957          \\
Gaussian  & \textbf{0.0404} & \textbf{0.0572} & 0.0918         \\ \hline
\end{tabular}
\caption{MSEs of methods with different priors (e.g., flow, EBM and Gaussian) in unsupervised image recovery.}
\label{tab:MSE_mask_recovery}
\end{table}
\begin{table}[h!]
\centering
\begin{tabular}{cccc}
\hline
\multicolumn{4}{c}{Salt and pepper mask}                     \\ \hline
Occ \%    & 30\%           & 50\%           & 70\%           \\ \hline
flow (ours)     & \textbf{46.2}  & \textbf{59.14} & \textbf{86.77} \\
EBM       & 52.78          & 61.91          & 88.27          \\
Gaussian  & 153.01         & 156.71         & 172.77         \\ \hline
\multicolumn{4}{c}{Single region mask}                       \\ \hline
mask size & $20\times20$          & $30\times30$          & $40\times40$          \\ \hline
flow (ours)      & \textbf{42.39} & \textbf{47.52} & \textbf{72.47} \\
EBM       & 49.16          & 51.59          & 77.39          \\
Gaussian  & 150.95         & 146.41         & 184.53         \\ \hline
\end{tabular}
\caption{FIDs of methods with different priors (e.g., flow, EBM and Gaussian) in unsupervised image recovery.}
\label{tab:FID_mask_recovery}
\end{table}

\section{Conclusion}

\paragraph{Summary} In this paper, we study modeling the latent space of data by the normalizing flow model and follow the philosophy of empirical Bayes to learn the latent space flow-based model (LFBM) from observed data. Specifically, we propose a novel top-down generative model with a latent space flow-based prior model. We propose a novel learning framework, in which the LFBM built on top of a top-down generative network serves as the prior model of the latent space, and is trained simultaneously with the top-down network in an MCMC-based MLE algorithm. We show that the learning algorithm with a short-run Langevin flow is a perturbation of MLE. The LFBM is more flexible and informative than Gaussian prior and less biased to MLE than the unnormalized energy-based prior (LEBM). We test our LFBM on a variety of tasks to validate its effectiveness.    
\paragraph{Limitation and Future Work} In contrast to EBM prior, the normalizing flow prior allows for efficient and unbiased sampling and training of the prior distribution. However, these good properties do not come for free, and the usage of the flow-based prior implies the assumption that the latent variables are computationally efficient to normalize and can be generated by a finite sequence of invertible transformations. This assumption might potentially limit the expressive power of the prior distribution for the latent space. Given the fact that EBM prior is an unnormalized probability density and doesn’t constrain the prior distribution by invertible functions, we may consider combining the representational flexibility of the EBM and the computational tractability of the normalizing flow model to tackle the above limitations. To be specific, we specify a latent space EBM (LEBM) and a latent space flow-based model (LFBM) together to represent the prior distribution. We  generalize the data space flow contrastive estimation (FCE) \cite{FlowEBM2020} to latent space for jointly training the LEBM and the LFBM without relying on MCMC. Flow contrastive estimation \cite{FlowEBM2020} can be considered as an improved version of noise contrastive estimation (NCE) \cite{GutmannH10} for training EBM, where the original Gaussian noise is replaced by a normalizing flow to make it closer to the target distribution and a stronger contrast to the EBM. Meanwhile, the flow model is updated by approximately minimizing the Jensen-Shannon divergence between the flow model and the target distribution. In our future work, we will learn both LFBM and LEBM together in the latent space via FCE.

\section*{Acknowledgments}
 The authors sincerely thank Dr. Ying Nian Wu at statistics department of  the University of California, Los Angeles (UCLA) for the helpful discussion on theoretical understanding of the learning algorithm. The authors would also like to thank the anonymous reviewers for providing constructive comments and suggestions to improve the work. Our work is also supported by XSEDE grant CIS210052.

% Use \bibliography{yourbibfile} instead or the References section will not appear in your paper
\bibliography{aaai23}

\begin{thebibliography}{45}
\providecommand{\natexlab}[1]{#1}

\bibitem[{An, Xie, and Li(2021)}]{AnXL21}
An, D.; Xie, J.; and Li, P. 2021.
\newblock Learning Deep Latent Variable Models by Short-Run {MCMC} Inference
  With Optimal Transport Correction.
\newblock In \emph{{IEEE} Conference on Computer Vision and Pattern Recognition
  (CVPR)}, 15415--15424.

\bibitem[{Chen et~al.(2017)Chen, Kingma, Salimans, Duan, Dhariwal, Schulman,
  Sutskever, and Abbeel}]{0022KSDDSSA17}
Chen, X.; Kingma, D.~P.; Salimans, T.; Duan, Y.; Dhariwal, P.; Schulman, J.;
  Sutskever, I.; and Abbeel, P. 2017.
\newblock Variational Lossy Autoencoder.
\newblock In \emph{International Conference on Learning Representations
  (ICLR)}.

\bibitem[{Ding and Gimpel(2021)}]{DingG21}
Ding, X.; and Gimpel, K. 2021.
\newblock FlowPrior: Learning Expressive Priors for Latent Variable Sentence
  Models.
\newblock In \emph{Proceedings of the 2021 Conference of the North American
  Chapter of the Association for Computational Linguistics: Human Language
  Technologies (NAACL-HLT)}, 3242--3258.

\bibitem[{Dinh, Krueger, and Bengio(2015)}]{DinhKB14}
Dinh, L.; Krueger, D.; and Bengio, Y. 2015.
\newblock {NICE:} Non-linear Independent Components Estimation.
\newblock In \emph{International Conference on Learning Representations (ICLR)
  Workshop}.

\bibitem[{Dinh, Sohl-Dickstein, and Bengio(2016)}]{dinh2016density}
Dinh, L.; Sohl-Dickstein, J.; and Bengio, S. 2016.
\newblock Density estimation using real nvp.
\newblock \emph{arXiv preprint arXiv:1605.08803}.

\bibitem[{Gao et~al.(2020)Gao, Nijkamp, Kingma, Xu, Dai, and Wu}]{FlowEBM2020}
Gao, R.; Nijkamp, E.; Kingma, D.~P.; Xu, Z.; Dai, A.~M.; and Wu, Y.~N. 2020.
\newblock Flow Contrastive Estimation of Energy-Based Models.
\newblock In \emph{IEEE Conference on Computer Vision and Pattern Recognition
  (CVPR)}, 7515--7525.

\bibitem[{Gutmann and Hyv{\"{a}}rinen(2010)}]{GutmannH10}
Gutmann, M.; and Hyv{\"{a}}rinen, A. 2010.
\newblock Noise-contrastive estimation: {A} new estimation principle for
  unnormalized statistical models.
\newblock In \emph{International Conference on Artificial Intelligence and
  Statistics (AISTATS)}, volume~9, 297--304.

\bibitem[{Han et~al.(2017)Han, Lu, Zhu, and Wu}]{HanLZW17}
Han, T.; Lu, Y.; Zhu, S.; and Wu, Y.~N. 2017.
\newblock Alternating Back-Propagation for Generator Network.
\newblock In \emph{{AAAI} Conference on Artificial Intelligence (AAAI)},
  1976--1984.

\bibitem[{Han et~al.(2020)Han, Nijkamp, Zhou, Pang, Zhu, and
  Wu}]{Han2020VAE-EBM}
Han, T.; Nijkamp, E.; Zhou, L.; Pang, B.; Zhu, S.-C.; and Wu, Y.~N. 2020.
\newblock Joint Training of Variational Auto-Encoder and Latent Energy-Based
  Model.
\newblock In \emph{Conference on Computer Vision and Pattern Recognition
  (CVPR)}.

\bibitem[{Heusel et~al.(2017)Heusel, Ramsauer, Unterthiner, Nessler, and
  Hochreiter}]{HeuselRUNH17}
Heusel, M.; Ramsauer, H.; Unterthiner, T.; Nessler, B.; and Hochreiter, S.
  2017.
\newblock GANs Trained by a Two Time-Scale Update Rule Converge to a Local Nash
  Equilibrium.
\newblock In \emph{Annual Conference on Neural Information Processing Systems
  (NIPS)}, 6626--6637.

\bibitem[{Huang et~al.(2017)Huang, Touati, Dinh, Drozdzal, Havaei, Charlin, and
  Courville}]{huang2017learnable}
Huang, C.-W.; Touati, A.; Dinh, L.; Drozdzal, M.; Havaei, M.; Charlin, L.; and
  Courville, A. 2017.
\newblock Learnable explicit density for continuous latent space and
  variational inference.
\newblock \emph{arXiv preprint arXiv:1710.02248}.

\bibitem[{Kingma and Ba(2015)}]{KingmaB14}
Kingma, D.~P.; and Ba, J. 2015.
\newblock Adam: {A} Method for Stochastic Optimization.
\newblock In \emph{3rd International Conference on Learning Representations
  (ICLR)}.

\bibitem[{Kingma and Dhariwal(2018)}]{KingmaD18}
Kingma, D.~P.; and Dhariwal, P. 2018.
\newblock Glow: Generative Flow with Invertible 1x1 Convolutions.
\newblock In \emph{Advances in Neural Information Processing Systems
  (NeurIPS)}, 10236--10245.

\bibitem[{Kingma et~al.(2016)Kingma, Salimans, Jozefowicz, Chen, Sutskever, and
  Welling}]{kingma2016improved}
Kingma, D.~P.; Salimans, T.; Jozefowicz, R.; Chen, X.; Sutskever, I.; and
  Welling, M. 2016.
\newblock Improved variational inference with inverse autoregressive flow.
\newblock \emph{Annual Conference on Neural Information Processing Systems
  (NIPS)}, 29: 4743--4751.

\bibitem[{Kingma and Welling(2014)}]{KingmaW13}
Kingma, D.~P.; and Welling, M. 2014.
\newblock Auto-Encoding Variational Bayes.
\newblock In \emph{International Conference on Learning Representations
  (ICLR)}.

\bibitem[{Krizhevsky, Hinton et~al.(2009)}]{krizhevsky2009learning}
Krizhevsky, A.; Hinton, G.; et~al. 2009.
\newblock Learning multiple layers of features from tiny images.

\bibitem[{Kumar et~al.(2020)Kumar, Babaeizadeh, Erhan, Finn, Levine, Dinh, and
  Kingma}]{KumarBEFLDK20}
Kumar, M.; Babaeizadeh, M.; Erhan, D.; Finn, C.; Levine, S.; Dinh, L.; and
  Kingma, D. 2020.
\newblock VideoFlow: {A} Conditional Flow-Based Model for Stochastic Video
  Generation.
\newblock In \emph{International Conference on Learning Representations
  (ICLR)}.

\bibitem[{Kumar et~al.(2019)Kumar, Ozair, Goyal, Courville, and
  Bengio}]{kumar2019maximum}
Kumar, R.; Ozair, S.; Goyal, A.; Courville, A.; and Bengio, Y. 2019.
\newblock Maximum entropy generators for energy-based models.
\newblock \emph{arXiv preprint arXiv:1901.08508}.

\bibitem[{LeCun et~al.(1998)LeCun, Bottou, Bengio, and Haffner.}]{mnist2010}
LeCun, Y.; Bottou, L.; Bengio, Y.; and Haffner., P. 1998.
\newblock Gradient-based learning applied to document recognition.
\newblock In \emph{Proceedings of the IEEE}, 2278--2324.

\bibitem[{Lee et~al.(2020)Lee, Liu, Wu, and Luo}]{CelebAMask-HQ}
Lee, C.-H.; Liu, Z.; Wu, L.; and Luo, P. 2020.
\newblock MaskGAN: Towards Diverse and Interactive Facial Image Manipulation.
\newblock In \emph{IEEE Conference on Computer Vision and Pattern Recognition
  (CVPR)}.

\bibitem[{Liu et~al.(2015)Liu, Luo, Wang, and Tang}]{LiuLWT15}
Liu, Z.; Luo, P.; Wang, X.; and Tang, X. 2015.
\newblock Deep Learning Face Attributes in the Wild.
\newblock In \emph{{IEEE} International Conference on Computer Vision (ICCV)},
  3730--3738.

\bibitem[{Ma, Zhou, and Hovy(2019)}]{MaZH19}
Ma, X.; Zhou, C.; and Hovy, E.~H. 2019.
\newblock {MAE:} Mutual Posterior-Divergence Regularization for Variational
  AutoEncoders.
\newblock In \emph{International Conference on Learning Representations
  (ICLR)}.

\bibitem[{Netzer et~al.(2011)Netzer, Wang, Coates, Bissacco, Wu, and
  Ng}]{netzer2011reading}
Netzer, Y.; Wang, T.; Coates, A.; Bissacco, A.; Wu, B.; and Ng, A.~Y. 2011.
\newblock Reading digits in natural images with unsupervised feature learning.
\newblock In \emph{NIPS Workshop on Deep Learning and Unsupervised Feature
  Learning}.

\bibitem[{Nijkamp et~al.(2019)Nijkamp, Hill, Zhu, and Wu}]{NijkampHZW19}
Nijkamp, E.; Hill, M.; Zhu, S.; and Wu, Y.~N. 2019.
\newblock Learning Non-Convergent Non-Persistent Short-Run {MCMC} Toward
  Energy-Based Model.
\newblock In \emph{Annual Conference on Neural Information Processing Systems
  (NeurIPS)}, 5233--5243.

\bibitem[{Nijkamp et~al.(2020)Nijkamp, Pang, Han, Zhou, Zhu, and
  Wu}]{NijkampP0ZZW20}
Nijkamp, E.; Pang, B.; Han, T.; Zhou, L.; Zhu, S.; and Wu, Y.~N. 2020.
\newblock Learning Multi-layer Latent Variable Model via Variational
  Optimization of Short Run {MCMC} for Approximate Inference.
\newblock In \emph{European Conference on Computer Vision (ECCV)}, volume
  12351, 361--378.

\bibitem[{Pang et~al.(2020)Pang, Han, Nijkamp, Zhu, and Wu}]{Pang0NZW20}
Pang, B.; Han, T.; Nijkamp, E.; Zhu, S.; and Wu, Y.~N. 2020.
\newblock Learning Latent Space Energy-Based Prior Model.
\newblock In \emph{Annual Conference on Neural Information Processing Systems
  (NeurIPS)}.

\bibitem[{Pang and Wu(2021)}]{PangW21}
Pang, B.; and Wu, Y.~N. 2021.
\newblock Latent Space Energy-Based Model of Symbol-Vector Coupling for Text
  Generation and Classification.
\newblock In \emph{International Conference on Machine Learning (ICML)}, volume
  139, 8359--8370.

\bibitem[{Pang et~al.(2021)Pang, Zhao, Xie, and Wu}]{PangZ0W21}
Pang, B.; Zhao, T.; Xie, X.; and Wu, Y.~N. 2021.
\newblock Trajectory Prediction With Latent Belief Energy-Based Model.
\newblock In \emph{{IEEE} Conference on Computer Vision and Pattern Recognition
  (CVPR)}, 11814--11824.

\bibitem[{Ping et~al.(2020)Ping, Peng, Zhao, and Song}]{PingPZ020}
Ping, W.; Peng, K.; Zhao, K.; and Song, Z. 2020.
\newblock WaveFlow: {A} Compact Flow-based Model for Raw Audio.
\newblock In \emph{International Conference on Machine Learning (ICML)}, volume
  119, 7706--7716.

\bibitem[{Rezende and Mohamed(2015)}]{RezendeM15}
Rezende, D.~J.; and Mohamed, S. 2015.
\newblock Variational Inference with Normalizing Flows.
\newblock In \emph{International Conference on Machine Learning (ICML)},
  volume~37, 1530--1538.

\bibitem[{Rubin and Thayer(1982)}]{rubin1982algorithms}
Rubin, D.~B.; and Thayer, D.~T. 1982.
\newblock EM algorithms for ML factor analysis.
\newblock \emph{Psychometrika}, 47(1): 69--76.

\bibitem[{Shi et~al.(2020)Shi, Xu, Zhu, Zhang, Zhang, and Tang}]{ShiXZZZT20}
Shi, C.; Xu, M.; Zhu, Z.; Zhang, W.; Zhang, M.; and Tang, J. 2020.
\newblock GraphAF: a Flow-based Autoregressive Model for Molecular Graph
  Generation.
\newblock In \emph{International Conference on Learning Representations
  (ICLR)}.

\bibitem[{Xiao, Yan, and Amit(2019)}]{xiao2019generative}
Xiao, Z.; Yan, Q.; and Amit, Y. 2019.
\newblock Generative latent flow.
\newblock \emph{arXiv preprint arXiv:1905.10485}.

\bibitem[{Xie et~al.(2019)Xie, Gao, Zheng, Zhu, and Wu}]{XieGZZW19}
Xie, J.; Gao, R.; Zheng, Z.; Zhu, S.; and Wu, Y.~N. 2019.
\newblock Learning Dynamic Generator Model by Alternating Back-Propagation
  through Time.
\newblock In \emph{{AAAI} Conference on Artificial Intelligence (AAAI)},
  5498--5507.

\bibitem[{Xie et~al.(2020{\natexlab{a}})Xie, Gao, Zheng, Zhu, and
  Wu}]{XieGZZW20}
Xie, J.; Gao, R.; Zheng, Z.; Zhu, S.; and Wu, Y.~N. 2020{\natexlab{a}}.
\newblock Motion-Based Generator Model: Unsupervised Disentanglement of
  Appearance, Trackable and Intrackable Motions in Dynamic Patterns.
\newblock In \emph{{AAAI} Conference on Artificial Intelligence (AAAI)},
  12442--12451.

\bibitem[{Xie et~al.(2020{\natexlab{b}})Xie, Lu, Gao, Zhu, and Wu}]{XieLGZW20}
Xie, J.; Lu, Y.; Gao, R.; Zhu, S.; and Wu, Y.~N. 2020{\natexlab{b}}.
\newblock Cooperative Training of Descriptor and Generator Networks.
\newblock \emph{{IEEE} Transactions on Pattern Analysis and Machine
  Intelligence}, 42(1): 27--45.

\bibitem[{Xie et~al.(2016)Xie, Lu, Zhu, and Wu}]{XieLZW16}
Xie, J.; Lu, Y.; Zhu, S.; and Wu, Y.~N. 2016.
\newblock A Theory of Generative ConvNet.
\newblock In \emph{International Conference on Machine Learning (ICML)},
  volume~48, 2635--2644.

\bibitem[{Xie et~al.(2022)Xie, Zhu, Li, and Li}]{coopflow}
Xie, J.; Zhu, Y.; Li, J.; and Li, P. 2022.
\newblock A Tale of Two Flows: Cooperative Learning of Langevin Flow and
  Normalizing Flow Toward Energy-Based Model.
\newblock In \emph{International Conference on Learning Representations
  (ICLR)}.

\bibitem[{Xing et~al.(2022)Xing, Gao, Han, Zhu, and Wu}]{XingGHZW22}
Xing, X.; Gao, R.; Han, T.; Zhu, S.; and Wu, Y.~N. 2022.
\newblock Deformable Generator Networks: Unsupervised Disentanglement of
  Appearance and Geometry.
\newblock \emph{IEEE Transactions on Pattern Analysis and Machine Intelligence
  (PAMI)}, 44(3): 1162--1179.

\bibitem[{Yang et~al.(2019)Yang, Huang, Hao, Liu, Belongie, and
  Hariharan}]{yang2019pointflow}
Yang, G.; Huang, X.; Hao, Z.; Liu, M.-Y.; Belongie, S.; and Hariharan, B. 2019.
\newblock Pointflow: 3d point cloud generation with continuous normalizing
  flows.
\newblock In \emph{International Conference on Computer Vision (ICCV)},
  4541--4550.

\bibitem[{Zenati et~al.(2018{\natexlab{a}})Zenati, Foo, Lecouat, Manek, and
  Chandrasekhar}]{zenati2018efficient}
Zenati, H.; Foo, C.~S.; Lecouat, B.; Manek, G.; and Chandrasekhar, V.~R.
  2018{\natexlab{a}}.
\newblock Efficient {GAN}-based anomaly detection.
\newblock \emph{arXiv preprint arXiv:1802.06222}.

\bibitem[{Zenati et~al.(2018{\natexlab{b}})Zenati, Foo, Lecouat, Manek, and
  Chandrasekhar}]{zenati2018anomaldetection}
Zenati, H.; Foo, C.~S.; Lecouat, B.; Manek, G.; and Chandrasekhar, V.~R.
  2018{\natexlab{b}}.
\newblock Efficient {GAN-based} Anomaly Detection.
\newblock \emph{arXiv: 1802.06222}.

\bibitem[{Zhang, Xie, and Barnes(2020)}]{ZhangXB20}
Zhang, J.; Xie, J.; and Barnes, N. 2020.
\newblock Learning Noise-Aware Encoder-Decoder from Noisy Labels by Alternating
  Back-Propagation for Saliency Detection.
\newblock In \emph{European Conference on Computer Vision (ECCV)}, volume
  12362, 349--366.

\bibitem[{Zhang et~al.(2021)Zhang, Xie, Barnes, and Li}]{nipsZhang21}
Zhang, J.; Xie, J.; Barnes, N.; and Li, P. 2021.
\newblock Learning Generative Vision Transformer with Energy-Based Latent Space
  for Saliency Prediction.
\newblock In \emph{Annual Conference on Neural Information Processing Systems
  (NeurIPS)}.

\bibitem[{Zhu et~al.(2019)Zhu, Xie, Liu, and Elgammal}]{ZhuXLE19}
Zhu, Y.; Xie, J.; Liu, B.; and Elgammal, A. 2019.
\newblock Learning Feature-to-Feature Translator by Alternating
  Back-Propagation for Generative Zero-Shot Learning.
\newblock In \emph{International Conference on Computer Vision (ICCV)},
  9843--9853.

\end{thebibliography}

\appendix
\section*{Appendix}

%\section{Theoretical Analysis}

\section{Maximizing Likelihood is Equivalent to Minimizing Kullback-Leibler Divergence}
\label{app:mle_kl}
We show that finding the maximum likelihood estimate of parameters $\theta$ in $p_{\theta}(x)$ amounts to minimizing the KL-divergence between the true data distribution $p_{\rm data}(x)$ and the model $p_{\theta}(x)$. We start with the objective that minimizes the KL-divergence between $p_{\rm data}(x)$ and $p_{\theta}(x)$ and derive the objective of the maximum likelihood estimator below,
\begin{eqnarray}
\begin{aligned}
 &\arg \min_{\theta} \mathbb{D}_{\text{KL}}(p_{\rm data}(x){\parallel}p_{\theta}(x)) \\
 =&\arg \min_{\theta} \mathbb{E}_{p_{\rm data}(x)} \left[\log \frac{p_{\rm data}(x)}{p_{\theta}(x)}\right]\\
=&\arg \min_{\theta}\mathbb{E}_{p_{\rm data}(x)} \left[\log p_{\rm data}(x) - \log p_{\theta}(x)\right],\\
=&\arg \min_{\theta} \mathbb{E}_{p_{\rm data}(x)} [\log p_{\rm data}(x)] - \mathbb{E}_{p_{\rm data}(x)} [\log p_{\theta}(x)],
\label{mle_mkl}
\end{aligned}
\end{eqnarray}
where the term $p_{\rm data}(x)$ is not dependent to the parameters $\theta$ and does not affect the argument of the minima in Eq.(\ref{mle_mkl}), thus we can omit this term and have 
\begin{eqnarray}
\begin{aligned}
 &\arg \min_{\theta} \mathbb{D}_{\text{KL}}(p_{\rm data}(x){\parallel}p_{\theta}(x)) \\
 =&\arg \min_{\theta}\mathbb{E}_{p_{\rm data}(x)} \left[ - \log p_{\theta}(x)\right]\\
=& \arg \max_{\theta} \mathbb{E}_{p_{\rm data}(x)} [\log p_{\theta}(x)].
\end{aligned}
\end{eqnarray}
Given a set of training examples $\{x_i\}_{i=1}^{N}\sim p_{\rm data}(x)$, according to the law of large numbers, the average over a large number of samples can approximate the expected value of that random variable. Formally, if $N$ goes to infinity,  
\begin{eqnarray}
\begin{aligned}
\lim_{N \rightarrow \infty}\frac{1}{N} \sum_{i=1}^{N} \log p_{\theta}(x_i) =  \mathbb{E}_{p_{\rm data}(x)} [\log p_{\theta}(x)],
\label{mle_mkl_2}
\end{aligned}
\end{eqnarray}
or in other words, if $N$ is large enough, the left term in Eq.(\ref{mle_mkl_2}), the data log-likelihood $\mathcal{L}(\theta|\{x_i\}_{i=1}^{N})$, can serve as an approximation of the right term. Thus,
\begin{eqnarray}
\arg \max_{\theta} \mathcal{L}(\theta|\{x_i\}_{i=1}^{N}) = \arg \min_{\theta} \mathbb{D}_{\text{KL}}(p_{\rm data}(x){\parallel}p_{\theta}(x)).
\end{eqnarray}
That is, MLE is equivalent to minimizing KL-divergence.

\section{Deriving the Gradient of the Likelihood}\label{app:derive_gradient}
Let $\theta=(\alpha, \beta)$ contain the parameters $\alpha$ of the prior model and the parameters $\beta$ of the top-down generation network. We present a detailed derivation of the maximum likelihood learning gradient as follows.
\begin{eqnarray}
\begin{aligned}
    \nabla_{\theta} \log p_{\theta}(x) 
    =& \frac{1}{p_{\theta}(x)} \nabla_{\theta} p_{\theta}(x) \\
    =& \frac{1}{p_{\theta}(x)} \nabla_{\theta} \left[\int p_{\theta}(x,z)dz \right] \\
    =& \frac{1}{p_{\theta}(x)} \left[\int  \nabla_{\theta} p_{\theta}(x,z)dz \right] \\
    =& \int \left[ \nabla_{\theta} p_{\theta}(x,z)\right] \frac{1}{p_{\theta}(x)}dz  \\
    =& \int \left[ \frac{1}{p_{\theta}(x,z)}\nabla_{\theta} p_{\theta}(x,z)\right] \frac{p_{\theta}(x,z)}{p_{\theta}(x)}dz  \\
     =& \int \left[ \nabla_{\theta} \log p_{\theta}(x,z)\right] p_{\theta}(z|x)dz  \\
    =& \mathbb{E}_{p_{\theta}(z|x)} \left[\nabla_{\theta} \log p_{\theta}(x,z)\right]  \\
    =& \mathbb{E}_{p_{\theta}(z|x)} \left[\nabla_{\theta} \log (p_{\alpha}(z)p_{\beta}(x|z))\right] \\
    =&\mathbb{E}_{p_{\theta}(z|x)}[\nabla_{\theta} (\log p_{\alpha}(z)+  \log p_{\beta}(x|z))] \\
    =&\mathbb{E}_{p_{\theta}(z|x)}[\nabla_{\theta} \log p_{\alpha}(z)+  \nabla_{\theta}\log p_{\beta}(x|z)] 
\end{aligned}
\end{eqnarray}
We can further separate the learning gradients of the prior model $\alpha$ and the top-down generation network $\beta$. That is,
\begin{align}
\nabla_{\alpha} \log p_{\theta}(x)=&\mathbb{E}_{p_{\theta}(z|x)}[\nabla_{\alpha} \log p_{\alpha}(z)],\\
\nabla_{\beta} \log p_{\theta}(x)=&\mathbb{E}_{p_{\theta}(z|x)}[\nabla_{\beta} \log p_{\beta}(x|z)].
\end{align}

\section{Implementation Details of the LFBM-VAE}

This section presents the implementation details of the baseline LFBM-VAE, which trains the latent space normalizing flow via variational inference, in which an inference network $q_{\phi}(z|x)$ is used to approximate the true posterior distribution $p_{\theta}(z|x)$. The inference network learns a direct mapping from observations to latent variables to mimic the intractable posterior. We formulate the inference model $q_{\phi}(z|x)$ via a flow-based model, which is of the form  
\begin{align}
&z_0 \sim q_{\phi_0}(z_0|x)=\mathcal{N}(z|\mu_{\phi_0}(x), \text{diag}(\sigma_{\phi_0}^2(x))),\\
 &z=f_{\phi_1}(z_0),  
\label{eq:prior_gen}
\end{align}
where $z_0$ follows a simple initial Gaussian distribution constructed via a bottom-up encoder with reparameterization trick, i.e., the mean function $\mu_{\phi_0}(x)$ and the standard-deviation function $\sigma_{\phi_0}(x)$ are specified by deep nets with parameters $\phi_0$. $f_{\phi_1}$ consists of a chain of invertible transformations that sequentially transform $z_0$ into a more complex distribution. Let $\phi=(\phi_0, \phi_1)$, where $\phi_0$ and $\phi_1$ are parameters of the encoder and the sequence of transformations, respectively. The resulting density of the flow-based inference model is given by
$q_{\phi}(z|x)=q_{\phi_0}(z_0|x)\prod_{l=1}^{L'}|\det (\frac{\partial z_{l-1}}{\partial z_l})|$, where $L'$ is the number of transformations in $f_{\phi_1}$ and $z_l$ denotes the output of the $l$-th transformation. Thus we have 
\begin{equation}
\log q_{\phi}(z|x)= \log q_{\phi_0}(z_0|x) + \sum_{l=1}^{L'}\log \left|\det \left(\frac{\partial z_{l-1}}{\partial z_l}\right)\right|.\notag
\label{eq:posterior_flow}
\end{equation}

Therefore, the learning objective of LFBM-VAE, $\mathcal{F}(\theta=(\alpha,\beta), \phi)$, is to maximize the log-likelihood of data while minimizing the KL divergence between the inference network and the posterior density, i.e.,
\begin{eqnarray}
\begin{aligned}
    \mathcal{F}(\theta, \phi)&=\log p_{\theta}(x)-\mathbb{D}_{\text{KL}}(q_{\phi}(z|x)||p_{\theta}(z|x)) \\
    &=\mathbb{E}_{q_{\phi}(z|x)}[\log p_{\beta}(x|z)]- \mathbb{D}_{\text{KL}}(q_{\phi}(z|x)||p_{\alpha}(z)) \\
    &=\mathbb{E}_{q_{\phi}(z|x)}[\log p_{\beta}(x|z)]-\mathbb{E}_{q_{\phi}(z|x)}[\log q_{\phi}(z|x)] \\
    &+ \mathbb{E}_{q_{\phi}(z|x)}[\log p_{\alpha}(z)]\\
    &=\mathbb{E}_{q_{\phi}(z|x)}[\log p_{\beta}(x|z)] + \mathbb{E}_{q_{\phi}(z|x)}[\log p_{\alpha}(z)]\\ 
    &-\mathbb{E}_{q_{\phi}(z|x)}[\log q_{\phi_0}(z_0|x)] \\
    &- \mathbb{E}_{q_{\phi}(z|x)}\left[\sum_{l=1}^{L'}\log \left|\det \left(\frac{\partial z_{l-1}}{\partial z_l}\right)\right|\right],
    \label{eq:vae_posterior_prior}
%    \end{split}
\end{aligned}
\end{eqnarray}
where $\mathbb{E}_{q_{\phi}(z|x)}[\cdot]$ is approximated by samples from the inference network $q_{\phi}(z|x)$. We update the parameters of prior network, generator network, and inference network via gradient ascent with  $\nabla_{\alpha}\mathcal{F}(\theta, \phi)$ $\nabla_{\beta}\mathcal{F}(\theta, \phi)$, and $\nabla_{\phi}\mathcal{F}(\theta, \phi)$ respectively. In practise, for the purpose of stable training, we might allow multiple steps of parameter updates for the prior and the generator at each iteration.

\section{Architectures and Hyperparameters}

\label{app:network_structure}
We show the generator network structure of both LFBM-MCMC and LFBM-VAE in Table~\ref{tab:generator structure}. The inference model structure of the LFBM-VAE is shown in Table~\ref{tab:inference structure}. For our normalizing flow prior model, we use a fully-connected version of GLOW model \cite{KingmaD18}. We by default use a 5-step normalizing flow prior model in all the experiments. Each step of flow is designed in \citet{KingmaD18}.  The baseline LFBM-VAE uses a flow-based inference network, and its architecture follows~\citet{kingma2016improved}. %The number of steps for each experiment is shown in Table~\ref{tab:inference structure}. 
Our key hyperparameters of LFBM-MCMC and LFBM-VAE for all experiments are shown in Table~\ref{tab:hyperparameters MCMC} and Table~\ref{tab:hyperparameters VAE}.

\begin{table*}[h!]
\begin{center}
\begin{tabular}{|c|c|c|}
\hline
% \multicolumn{3}{|c|}{\textbf{Flow-based Model}}              \\ \hline
% Layers & In-Out Size & Stride \\ \hline
%       \\\hline \hline
\multicolumn{3}{|c|}{\textbf{Generator Model for SVHN}, $n_g=64$} \\ \hline
\multicolumn{1}{|c|}{Layers} & In-Out Size & Stride \\  \hline
Input  &    $1\times1\times100$   &        \\ 
$4 \times4$ convT($n_g \times 8$), LReLU  &     $4 \times4\times(n_g \times 8)$       & 1        \\ 
$4 \times4$ convT($n_g \times 4$), LReLU  &     $8 \times8\times(n_g \times 4)$       & 2        \\ 
$4 \times4$ convT($n_g \times 2$), LReLU  &     $16 \times16\times(n_g \times 2)$       & 2        \\ 
$4 \times4$ convT($3$), Tanh  &     $32 \times32\times3$       & 2         \\\hline \hline

\multicolumn{3}{|c|}{\textbf{Generator Model for CIFAR-10}, $n_g=128$}              \\ \hline
\multicolumn{1}{|c|}{Layers} & In-Out Size & Stride      \\\hline 
Input   &  $1\times1\times128$     &        \\  
$8 \times8$ convT($n_g \times 8$), LReLU  &     $8 \times8\times(n_g \times 8)$       & 1        \\ 
$4 \times4$ convT($n_g \times 4$), LReLU  &     $16 \times16\times(n_g \times 4)$       & 2        \\ 
$4 \times4$ convT($n_g \times 2$), LReLU  &     $32 \times32\times(n_g \times 2)$       & 2        \\ 
$3 \times3$ convT($3$), Tanh  &     $32 \times32\times3$       & 1         \\\hline \hline

\multicolumn{3}{|c|}{\textbf{Generator Model for CelebA}, $n_g=128$}              \\ \hline
\multicolumn{1}{|c|}{Layers} & In-Out Size & Stride \\ \hline
Input   &  $1\times1\times100$     &        \\ 
$4 \times4$ convT($n_g \times 8$), LReLU  &     $4 \times4\times(n_g \times 8)$       & 1        \\ 
$4 \times4$ convT($n_g \times 4$), LReLU  &     $8 \times8\times(n_g \times 4)$       & 2        \\ 
$4 \times4$ convT($n_g \times 2$), LReLU  &     $16 \times16\times(n_g \times 2)$       & 2        \\ 
$4 \times4$ convT($n_g \times 1$), LReLU  &     $32 \times32\times(n_g \times 1)$       & 2       \\ 
$4 \times4$ convT($3$), Tanh                & $64 \times64 \times 3$   & 2      \\\hline                         
\end{tabular} 
\end{center}
\caption{Generator model architectures for datasets SVHN, CIFAR-10, and CelebA. }
\label{tab:generator structure}
\end{table*}

\begin{table*}[h!]
\begin{center}
\begin{tabular}{|c|c|c|}
\hline

\multicolumn{3}{|c|}{\textbf{Inference Model for SVHN}, $n_i=10$} \\ \hline
\multicolumn{1}{|c|}{Layers} & In-Out Size & Stride \\  \hline
Input  &    $32\times32\times3$   &     \\ 
$3 \times3$ conv($n_i$), LReLU  &     $32 \times32\times(n_i)$       & 1        \\ 
$4 \times4$ conv($n_i \times 2$), LReLU  &     $16 \times16\times(n_i \times 2)$       & 2        \\ 
$4 \times4$ conv($n_i \times 4$), LReLU  &     $8 \times8\times(n_i \times 4)$       & 2        \\
$4 \times4$ conv($n_i \times 8$), LReLU  &     $4 \times4\times(n_i \times 8)$       & 2        \\
$4 \times4$ conv(100) &     $1 \times1\times100$       & 1 \\\hline
\multicolumn{3}{|c|}{20-step inverse autoregressive flow with a hidden size of 200} \\
\hline \hline

\multicolumn{3}{|c|}{\textbf{Inference Model for CIFAR-10}, $n_i=64$} \\ \hline
\multicolumn{1}{|c|}{Layers} & In-Out Size & Stride \\  \hline
Input  &    $32\times32\times3$   &     \\ 
$3 \times3$ conv($n_i$), LReLU  &     $32 \times32\times(n_i)$       & 1        \\ 
$4 \times4$ conv($n_i \times 2$), LReLU  &     $16 \times16\times(n_i \times 2)$       & 2        \\ 
$4 \times4$ conv($n_i \times 4$), LReLU  &     $8 \times8\times(n_i \times 4)$       & 2        \\
$4 \times4$ conv($n_i \times 8$), LReLU  &     $4 \times4\times(n_i \times 8)$       & 2        \\
$4 \times4$ conv(128) &     $1 \times1\times128$       & 1 \\\hline
\multicolumn{3}{|c|}{5-step inverse autoregressive flow with a hidden size of 256} \\
\hline \hline

\multicolumn{3}{|c|}{\textbf{Inference Model for CelebA}, $n_i=64$} \\ \hline
\multicolumn{1}{|c|}{Layers} & In-Out Size & Stride \\  \hline
Input  &    $64\times64\times3$   &     \\ 
$4 \times4$ conv($n_i$), LReLU  &     $32 \times32\times(n_i)$       & 2        \\ 
$4 \times4$ conv($n_i \times 2$), LReLU  &     $16 \times16\times(n_i \times 2)$       & 2        \\ 
$4 \times4$ conv($n_i \times 4$), LReLU  &     $8 \times8\times(n_i \times 4)$       & 2        \\
$4 \times4$ conv($n_i \times 4$), LReLU  &     $4 \times4\times(n_i \times 4)$       & 2        \\
$4 \times4$ conv(100) &     $1 \times1\times100$       & 1 \\\hline
\multicolumn{3}{|c|}{5-step inverse autoregressive flow with a hidden size of 200} \\
\hline                         
\end{tabular} 
\end{center}
\caption{Inference model architectures for datasets SVHN, CIFAR-10, and CelebA. }
\label{tab:inference structure}
\end{table*}

\begin{table*}[h!]
\begin{center}
\begin{tabular}{|c|c|c|c|}
\hline
\textbf{Parameter} & \textbf{SVHN} & \textbf{CIFAR-10} & \textbf{CelebA} \\ \hline
number of MCMC steps in training & 20 & 40 & 20 \\ 
number of MCMC steps in testing & 400 & 800 & 400 \\ 
Langevin step size & 0.1 & 0.1 & 0.1\\ 
image size & 32 & 32 & 64\\ 
batch size &100  & 100 & 100\\ 
latent space dimension & 100 & 128 & 100\\
standard deviation of the residual in the generator & 1.0 & 1.0 & 1.0\\ 
learning rate of generator & 0.0004 & 0.00038 & 0.0003\\ 
learning rate of flow-based model & 0.0004  & 0.00038 & 0.0003\\ 
learning decay of generator & 0.998 & 0.998 & 0.998\\ 
learning decay of flow-based model & 0.998  & 0.998  & 0.998\\ \hline 
\end{tabular} 
\end{center}
\caption{Key hyperparameters used in different experiments of LFBM-MCMC}
\label{tab:hyperparameters MCMC}
\end{table*}

\begin{table*}[h!]
\begin{center}
\begin{tabular}{| c | c| c | c |}
\hline
\textbf{Parameter} & \textbf{SVHN} & \textbf{CIFAR-10} & \textbf{CelebA} \\ \hline
image size & 32 & 32 & 64\\ 
batch size & 100  & 100 & 256\\ 
latent space dimension & 100 & 128 & 100\\
standard deviation of the residual in the generator & 0.5 & 0.25 & 0.25\\ 
learning rate of generator & 0.008 & 0.002 & 0.002\\ 
learning rate of flow-based model (prior) & 0.0006  & 0.0012 & 0.00015\\ 
learning rate of inference model & 0.0004  & 0.0002 & 0.0001\\ 
learning decay of generator & 0.99 & 0.99 & 0.99\\ 
learning decay of flow-based model (prior) & 0.99  & 0.99  & 0.99\\ 
learning decay of inference model  & 0.99  & 0.99  & 0.99\\
number of update steps for the prior and inference models at an iteration & 1 & 6 & 6 \\  
\hline 
\end{tabular} 
\end{center}
\caption{Key hyperparameters used in different experiments of LFBM-VAE}
\label{tab:hyperparameters VAE}
\end{table*}

\section{More Qualitative Results for Image Generation and Reconstruction }
\begin{figure*}[h!]
\centering
\begin{subfigure}{.3\linewidth}
    \centering
    \includegraphics[width=0.95\textwidth]{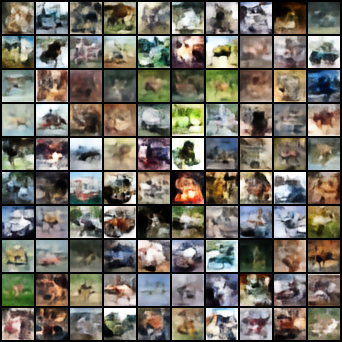}
    \caption{CIFAR10}
\end{subfigure}
    \hspace{-0.5em}
\begin{subfigure}{.3\linewidth}
    \centering
    \includegraphics[width=0.95\textwidth]{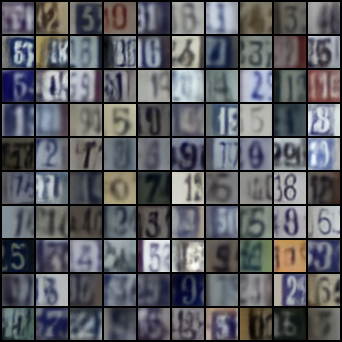}
    \caption{SVHN}
\end{subfigure}
  \hspace{-0.5em}
\begin{subfigure}{.3\linewidth}
    \centering
    \includegraphics[width=0.95\textwidth]{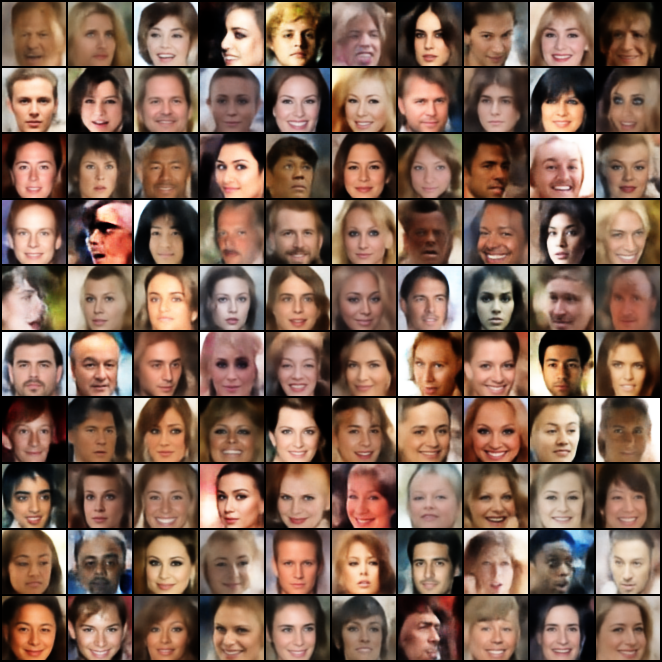}
    \caption{CelebA}
\end{subfigure}
\caption{Generated examples from the baseline LFBM-VAE models trained on the CIFAR10 (32$\times$ 32), SVHN, (32$\times$ 32) and CelebA (64$\times$ 64) datasets. The LFBM-VAE model is trained with an inference model for amortized inference.} \label{fig:sample_vae}
\end{figure*}

\begin{figure*}[h!]
\centering
\begin{subfigure}{.27\linewidth}
    \centering
    \includegraphics[width=0.95\textwidth]{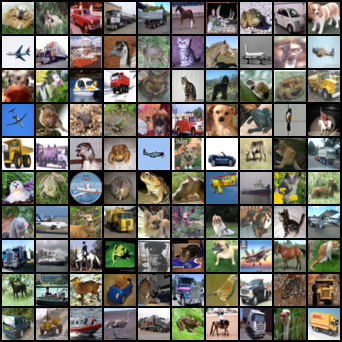}
\end{subfigure}
    \hspace{-0.5em}
\begin{subfigure}{.27\linewidth}
    \centering
    \includegraphics[width=0.95\textwidth]{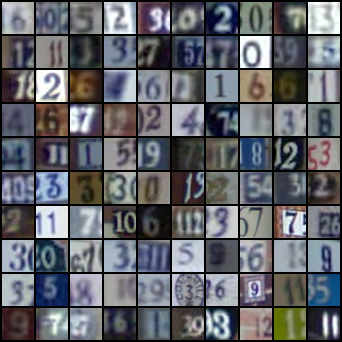}
\end{subfigure}
  \hspace{-0.5em}
\begin{subfigure}{.27\linewidth}
    \centering
    \includegraphics[width=0.95\textwidth]{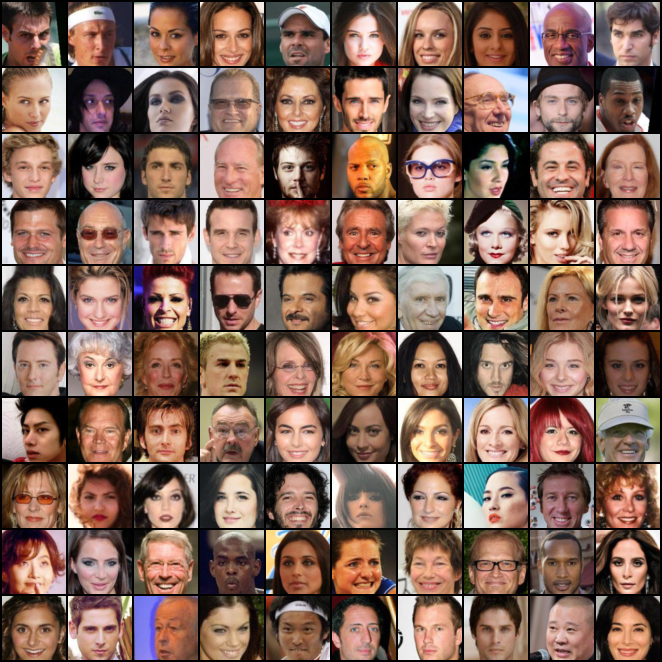}
\end{subfigure}
\vfill \vspace{2mm}
\begin{subfigure}{.27\linewidth}
    \centering
    \includegraphics[width=0.95\textwidth]{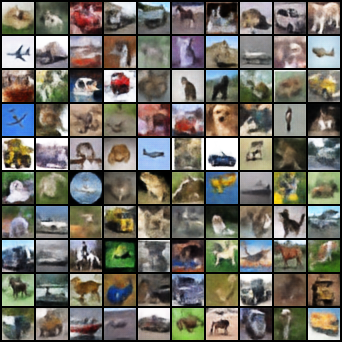}
    \caption{CIFAR10}
\end{subfigure}
    \hspace{-0.5em}
\begin{subfigure}{.27\linewidth}
    \centering
    \includegraphics[width=0.95\textwidth]{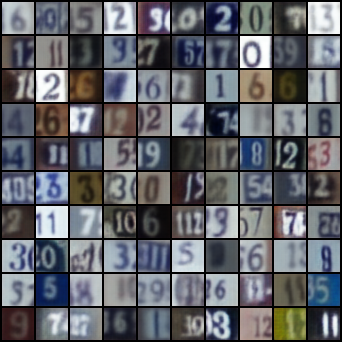}
    \caption{SVHN}
\end{subfigure}
 \hspace{-0.5em}
\begin{subfigure}{.27\linewidth}
    \centering
    \includegraphics[width=0.95\textwidth]{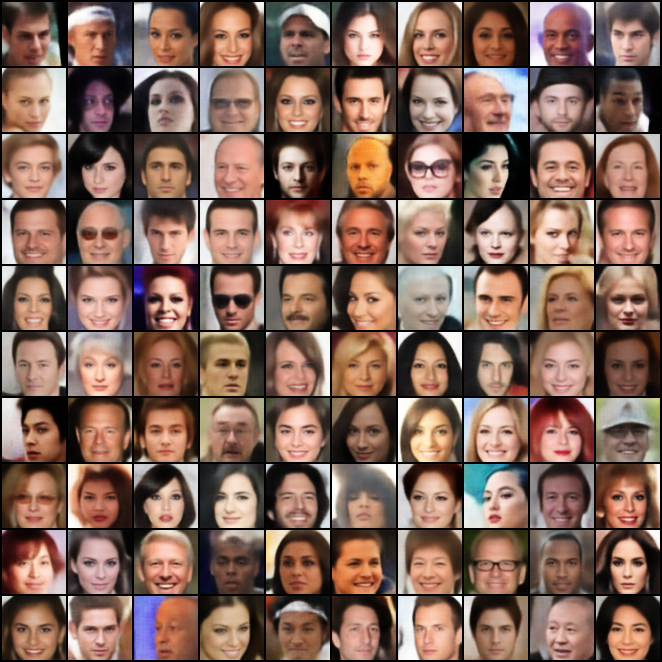}
    \caption{CelebA}
\end{subfigure}
\caption{Image reconstruction results by the LFBM-MCMC trained with short-run Langevin flow inference on CIFAR10 (32$\times$ 32), SVHN, (32$\times$ 32) and CelebA (64$\times$ 64) datasets. Top row: observed images; Bottom row: reconstructed images.} 
\label{fig:recons_mcmc}
\end{figure*}

\begin{figure*}[h!]
\centering
\begin{subfigure}{.27\linewidth}
    \centering
    \includegraphics[width=0.95\textwidth]{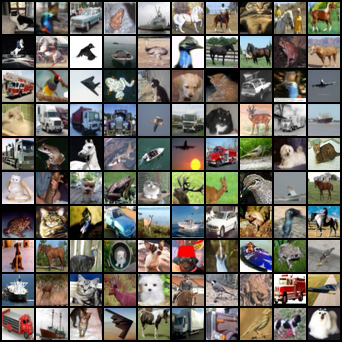}
\end{subfigure}
    \hspace{-0.5em}
\begin{subfigure}{.27\linewidth}
    \centering
    \includegraphics[width=0.95\textwidth]{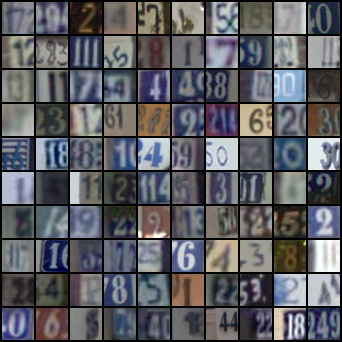}
\end{subfigure}
  \hspace{-0.5em}
\begin{subfigure}{.27\linewidth}
    \centering
    \includegraphics[width=0.95\textwidth]{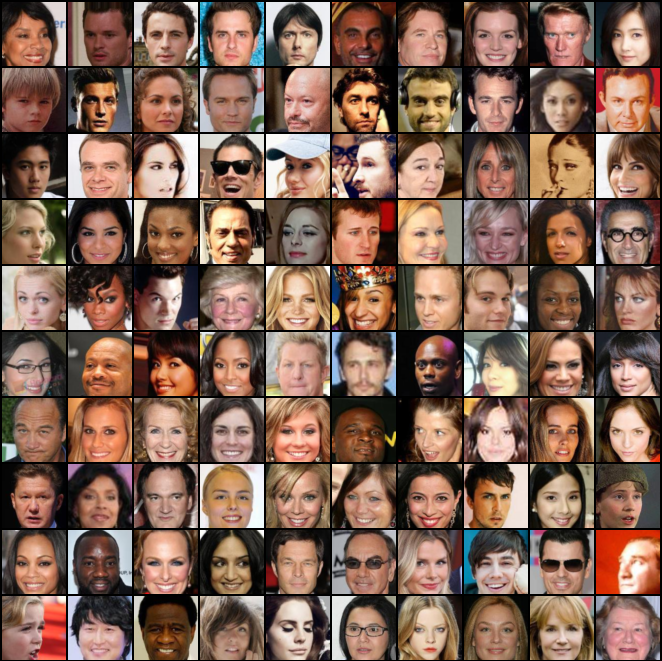}
\end{subfigure}
%\vspace{1.0cm} 
\vfill \vspace{2mm}
\begin{subfigure}{.27\linewidth}
    \centering
    \includegraphics[width=0.95\textwidth]{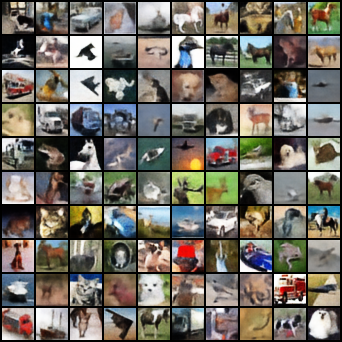}
    \caption{CIFAR10}
\end{subfigure}
    \hspace{-0.5em}
\begin{subfigure}{.27\linewidth}
    \centering
    \includegraphics[width=0.95\textwidth]{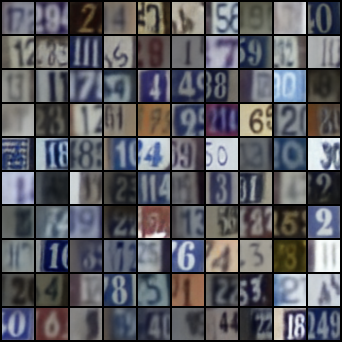}
    \caption{SVHN}
\end{subfigure}
 \hspace{-0.5em}
\begin{subfigure}{.27\linewidth}
    \centering
    \includegraphics[width=0.95\textwidth]{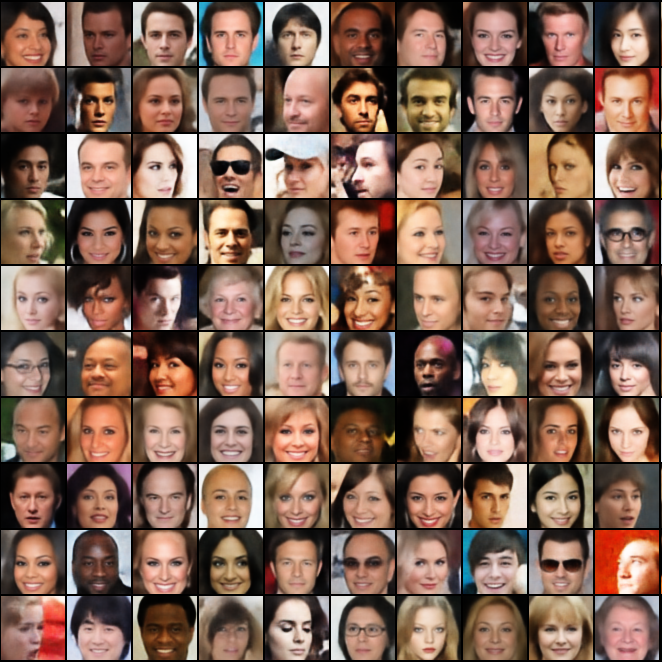}
    \caption{CelebA}
\end{subfigure}
\caption{Image reconstruction results by the baseline model, LFBM-VAE, trained with amortized inference on CIFAR10 (32$\times$ 32), SVHN, (32$\times$ 32) and CelebA (64$\times$ 64) datasets. The top row: observed images; the bottom row: reconstructed images. } 
\label{fig:recons_vae}
\end{figure*}

For the purpose of qualitative comparison, we show the generated examples by the baseline LFBM-VAE that uses amortized inference in Figure~\ref{fig:sample_vae}. As to image reconstruction task, we show the reconstructed images of our LFBM-MCMC and the baseline LFBM-VAE in Figure~\ref{fig:recons_mcmc} and Figure~\ref{fig:recons_vae} respectively. For a fair comparison, both LFBM-MCMC and LFBM-VAE use the same design of network architectures for the top-down generator and the normalizing flow prior, except that the baseline LFBM-VAE has an extra inference network for the purpose of amortized inference.

\section{Computational Time}
In Table~\ref{tab:time}, we compare our LFBM-MCMC with other baseline approaches, e.g., LFBM-VAE, LEBM and Gaussian prior model in terms of computational time for training and image generation. All the experiments are carried on the CelebA dataset using a PC with a single NVIDIA RTX A6000 GPU. We use a batch size of 256 images for all the models to make a fair comparison. We can see that the baseline LFBM-VAE is much faster than the other three MCMC-based models in the training stage because of the usage of the amortized inference. Unsurprisingly, our LFBM-MCMC has the longest training time since it has an MCMC-based inference and needs to update the parameters of the latent space normalizing flow model. As to the image generation, we report the computational time for generating one batch of images. We can see that both LFBM-MCMC and LFBM-VAE models have almost the same computational time to synthesize images, and are slightly faster than the LEBM model, because the latter needs an iterative MCMC process to sample from the energy-based prior distribution when generating examples. The Gaussian prior model is the fastest since its prior distribution is the simplest. %Table~\ref{tab:model_size} also presents a comparison of model size between the LFBM and LFBM-VAE in terms of number of model parameters. Since the LFBM doesn't have an extra inference network for amortized inference, its model size is much smaller than that of the LFBM-VAE.  

\begin{table*}[h!]

\begin{center}
%\begin{small}
\begin{tabular}{|c|cccc|}
\hline
Model & LFBM-MCMC (ours) & LFBM-VAE &  LEBM & Gaussian         \\ \hline
training (minutes / epoch) & 20.7 & 5.5 &  14.8 & 14.3 \\ \hline
synthesizing (milliseconds / batch) & 105 & 104 &  116 & 66 \\ \hline
%Inference time per batch &  &  &  &  \\ \hline
\end{tabular}
%\end{small}
\end{center}
\caption{Comparison among LFBM-MCMC, LFBM-VAE, LEBM, and Gaussian prior in terms of computational time for model training and image generation. }
\label{tab:time}
\end{table*}

\section{High Resolution Image Generation}
\label{sec:celeba256}
We further demonstrate the power of the LFBM-MCMC by learning the model on the 256 $\times$ 256 version of CelebAMask-HQ \cite{CelebAMask-HQ} dataset that contains 30,000 high-resolution face images. In our experiments, we only use the original face images as training examples and do not adopt any other information provided by the dataset, e.g., labels or masks. We compare Gaussian prior model \cite{HanLZW17}, LEBM \cite{Pang0NZW20}, LFBM-VAE and our LFBM-MCMC using the same generator structure shown in Table \ref{tab:generator structure celeba256}. This is a simple top-down ConvNet-based generator for $256 \times 256$ images because our purpose is to demonstrate the advantage of our model in a fair comparison with other baselines. For the Gaussian prior model, LEBM and LFBM-MCMC, we use a 60-step Langevin flow for approximate inference. For the LFBM-VAE, we use a flow-based inference network, which is a symmetric structure of the generator network, followed by a 5-step autoregressive flow with a hidden size of 256. %. 
 For the LFBM and LFBM-VAE, we use a 10-step normalizing flow prior model with a hidden size of 128. As to the LEBM, we use the EBM prior structure for generating $128 \times 128$ resolution CelebA images proposed by \cite{Pang0NZW20}, i.e., a 3-layer fully connected neural network with a hidden size of 1,000, which has about twice as many parameters as our 10-step flow-based prior model.

\begin{table}[htb!]
\begin{center}
\begin{tabular}{|c|c|c|}
\hline
\multicolumn{1}{|c|}{Layers} & In-Out Size & Stride \\ \hline
Input   &  $1\times1\times128$     &        \\ 
$4 \times4$ convT(2048), LReLU  &     $4 \times4\times2048$       & 1        \\ 
$4 \times4$ convT(1024), LReLU  &     $8 \times8\times1024$       & 2        \\ 
$4 \times4$ convT(512), LReLU  &     $16 \times16\times512$       & 2        \\ 
$4 \times4$ convT(256), LReLU  &     $32 \times32\times256$       & 2       \\ 
$4 \times4$ convT(256), LReLU  & $64 \times64 \times 256$   & 2  \\    
$4 \times4$ convT(128), LReLU  & $128 \times128 \times 128$   & 2  \\
$4 \times4$ convT(3), LReLU  & $256 \times256 \times 3$   & 2 \\\hline                     
\end{tabular} 
\end{center}
\caption{Generator model architecture for the high-resolution ($256 \times 256$) CelebAMask-HQ dataset.}
\label{tab:generator structure celeba256}
\end{table}

\begin{figure*}[h!]
\centering
\begin{subfigure}{.49\linewidth}
    \centering
    \includegraphics[width=0.95\textwidth]{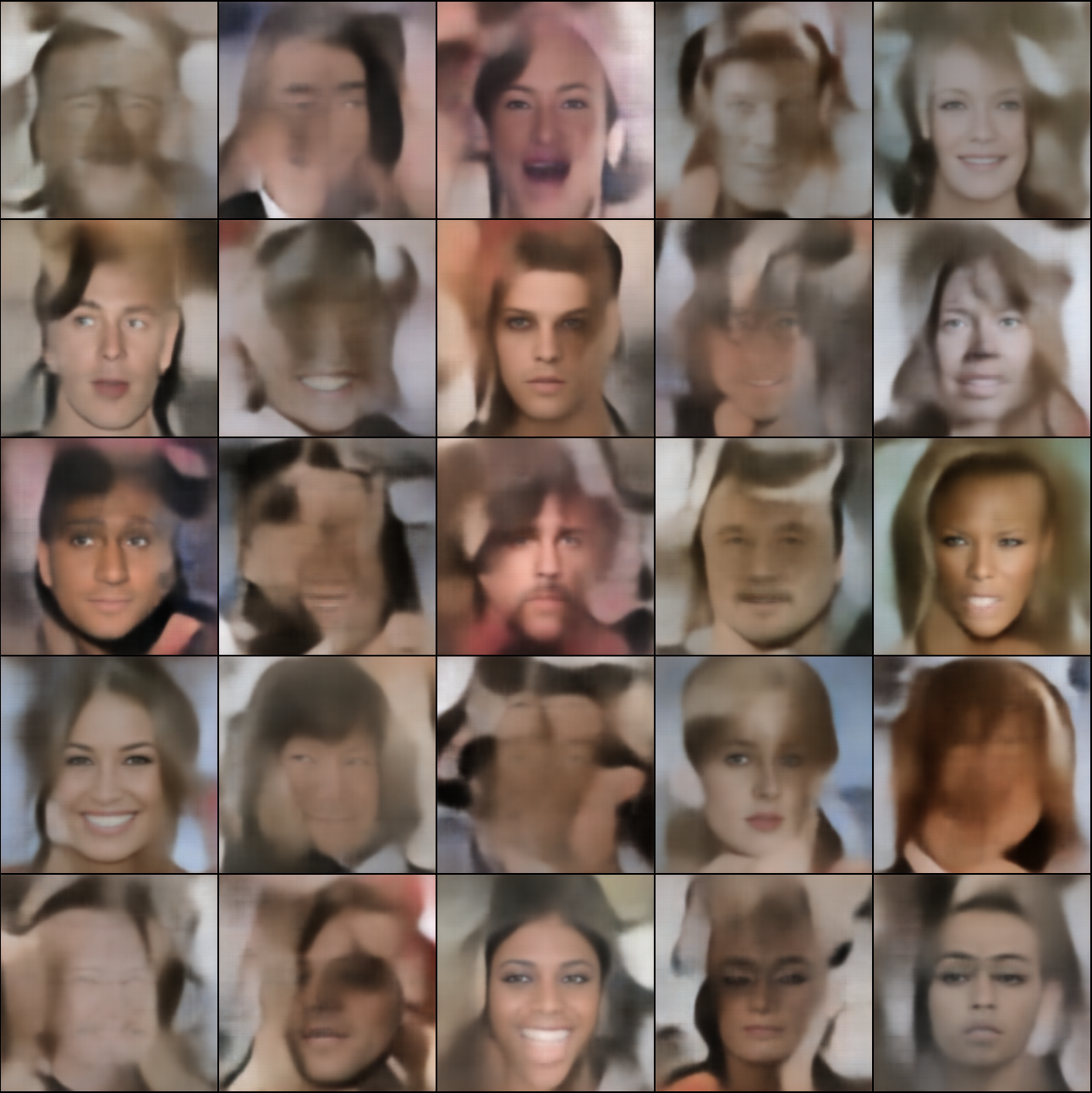}
    \caption{Gaussian prior}
\end{subfigure}
 \hspace{-0.5em}
\begin{subfigure}{.49\linewidth}
    \centering
    \includegraphics[width=0.95\textwidth]{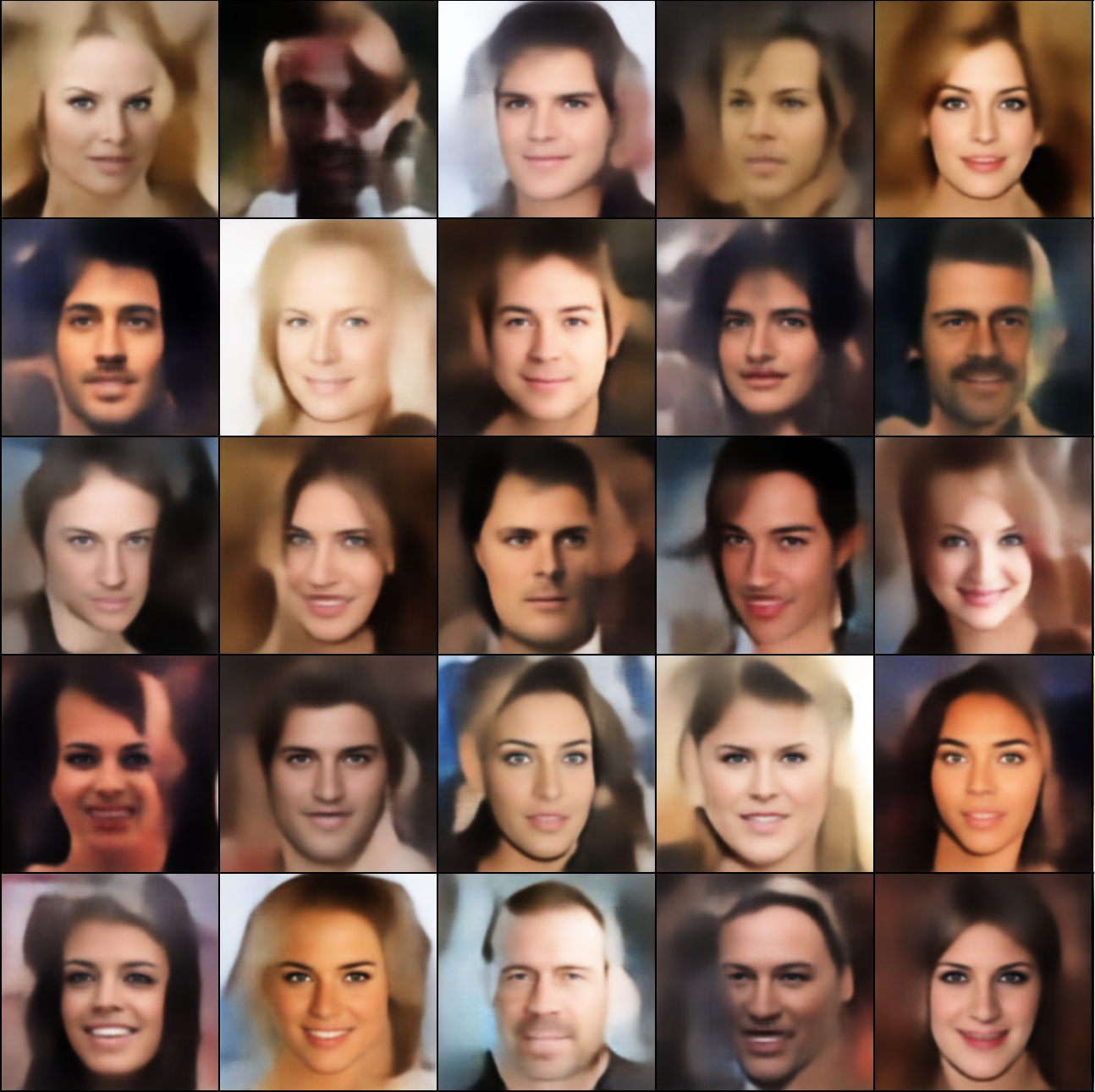}
    \caption{LEBM}
\end{subfigure}
%\vspace{1.0cm} 
\vfill
\begin{subfigure}{.49\linewidth}
    \centering
    \includegraphics[width=0.95\textwidth]{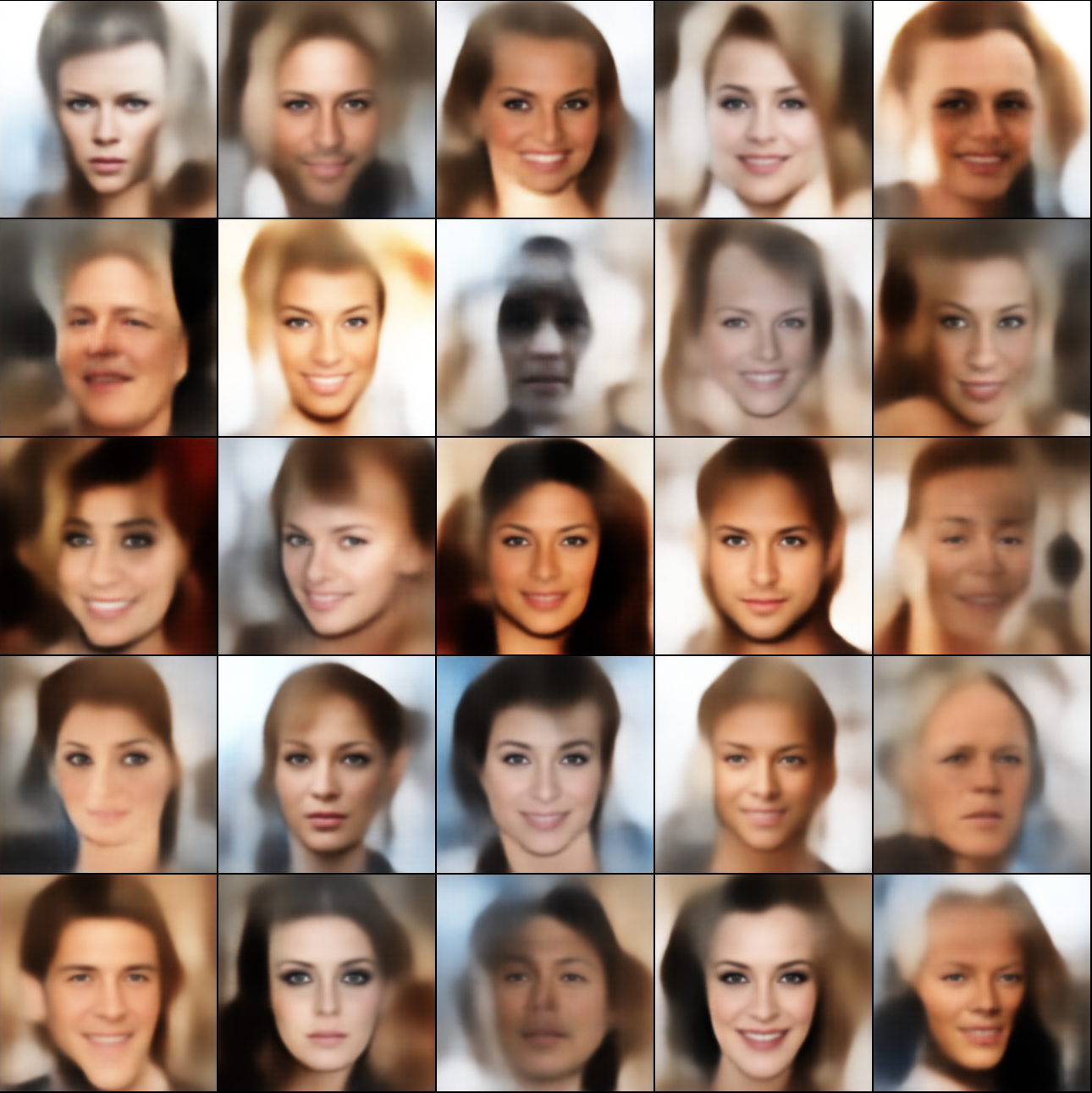}
    \caption{LFBM-VAE}
\end{subfigure}
 \hspace{-0.5em}
\begin{subfigure}{.49\linewidth}
    \centering
    \includegraphics[width=0.95\textwidth]{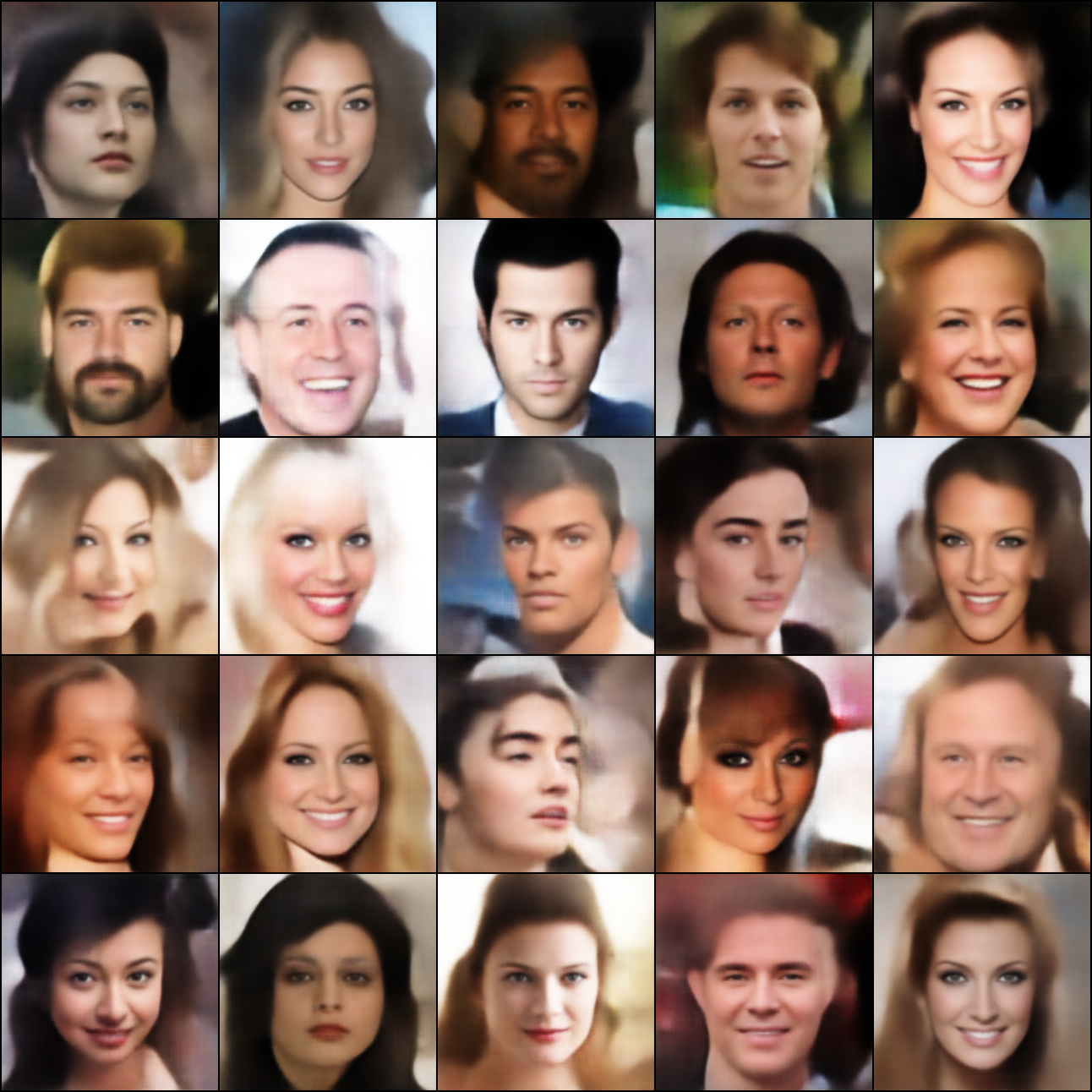}
    \caption{LFBM-MCMC (ours)}
\end{subfigure}
\caption{Generated examples by the models trained on the high-resolution CelebAMask-HQ (256 $\times$ 256) dataset.}
\label{fig:celeba256}
\end{figure*}

We show quantitative results in terms of FID and model size in Table \ref{tab:celeba256} and qualitative results in Figure \ref{fig:celeba256}. We can see that our LFBM-MCMC model trained with short-run Langevin flow outperforms all other baseline models by large margins in terms of FID. The relatively high FID of the LFBM-VAE indicates  the challenge of designing an optimal inference network structure that can approximate the posterior distribution. This further justifies the advantage of the design-free Langevin flow inference engine in our proposed LFBM-MCMC framework.

\begin{table*}[h!]
\begin{center}
%\begin{small}
%\begin{tabular}{m{6em}|m{8em}m{9em}m{6em}m{6em}}
\begin{tabular}{|c|cccc|}
\hline
Model & Prior network size & Inference network size &  Generator size & FID $\downarrow$        \\ \hline
Gaussian Prior & $-$ & $-$ &  49,817,731 & 102.61 \\ \hline
LEBM & 1,131,001 & $-$ & 49,817,731 & 94.03 \\ \hline
LFBM-VAE & 586,240 & 59,194,240 & 49,817,731 & 129.65 \\ \hline
LFBM-MCMC (ours) & 586,240 & $-$ & 49,817,731 & 75.20 \\\hline
\end{tabular}
%\end{small}
\end{center}
\caption{Sizes and FID scores for different models on CelebAMask-HQ (256 $\times$ 256) dataset.}
\label{tab:celeba256}
\end{table*}

\section{Comparison with the CoopFlow \cite{coopflow} Algorithm}

In this section, we discuss the differences between our LFBM-MCMC framework and the CoopFlow framework \cite{coopflow}. Even though they have similar components, e.g., a short-run Langevin flow and a normalizing flow, two frameworks differ in the following aspects:
\begin{itemize}
    \item \textbf{Normalizing flow}. The normalizing flow in our LFBM-MCMC framework is built on the low-dimensional latent space of the data, which need to be inferred with a top-down generator model. On the contrary, the normalizing flow in the CoopFlow framework is built on the high-dimensional observable data space.

\item \textbf{Langevin flow}. The Langevin dynamics used in our LFBM-MCMC framework is intended to draw samples from the posterior distribution $p(z|x)$, which is derived from the prior distribution and the generator according to the Bayes rule, while the Langevin dynamics in the CoopFlow is used to draw samples from a specified energy-based model $p(x)$. The former is a low-dimensional Langevin flow for inference purpose, while the latter is a high-dimensional Langevin flow for synthesis purpose.

\item  \textbf{Relationship between Langevin flow and normalizing flow}. In our LFBM-MCMC framework, the normalizing flow serves as a prior distribution of the latent variables, while the Langevin flow serves as an approximate posterior distribution of the latent variable. The posterior distribution is dependent on the parameters of the prior distribution. That is, once the normalizing flow prior model is updated, the Langevin flow inference model is also updated automatically at the same time. However, in the CoopFlow, the normalizing flow serves as an initializer of the Langevin flow that draws samples from the energy-based model, or in other words, the normalizing flow is an amortized sampler of the Langevin flow. The parameters of the Langevin flow and those of the normalizing flow in the CoopFlow model are independent.

\item \textbf{Training stage}. Even though both learning algorithms are based on maximum likelihood, they belong to different learning schemes. The CoopFlow follows the philosophy of cooperative learning via MCMC teaching \cite{XieLGZW20}, while our LFBM-MCMC framework follows the philosophy of empirical Bayes.

\item \textbf{Testing stage}. The resulting model in our LFBM-MCMC framework is a composition of the normalizing flow and the top-down generator network, while the resulting model in the CoopFlow is a composition of the normalizing flow and the Langevin~flow.
\end{itemize}

\bigskip
\end{document}